\documentclass[letterpaper]{article}
\usepackage{uai2019}
\usepackage[margin=1in]{geometry}

\usepackage{times}

\usepackage{amsmath,amssymb,amsthm}
\usepackage{graphicx}
\usepackage{algorithm}
\usepackage{algorithmic}
\usepackage{float}
\usepackage[round,sort]{natbib}

\usepackage{scalerel,stackengine}
\stackMath
\newcommand\reallywidehat[1]{%
\savestack{\tmpbox}{\stretchto{%
  \scaleto{%
    \scalerel*[\widthof{\ensuremath{#1}}]{\kern-.6pt\bigwedge\kern-.6pt}%
    {\rule[-\textheight/2]{1ex}{\textheight}}%WIDTH-LIMITED BIG WEDGE
  }{\textheight}% 
}{0.5ex}}%
\stackon[1pt]{#1}{\tmpbox}%
}

\newtheorem{theorem}{Theorem}

\newtheorem{lemma}[theorem]{Lemma}

\newtheorem{remark}{Remark}

\renewcommand{\vec}[1]{\boldsymbol{#1}}
\newcommand{\set}[1]{\mathcal{#1}}
\newcommand{\op}[1]{\boldsymbol{#1}}

\DeclareMathOperator{\R}{\mathbb{R}}

\DeclareMathOperator{\E}{\mathbb{E}}

\DeclareMathOperator*{\argmax}{arg\,max}
\DeclareMathOperator*{\argmin}{arg\,min}

\title{Gap-Increasing Policy Evaluation for\\ Efficient and Noise-Tolerant Reinforcement Learning}

\author{
\bf{Tadashi Kozuno}\\
Neural Computation Unit\\
Okinawa Inst. of Sci. and Tech.\\
Okinawa, Japan\\
\texttt{tadashi.kozuno@oist.jp}
\And
\bf{Dongqi Han}\\
Cognitive Neurorobotics Research Unit\\
Okinawa Inst. of Sci. and Tech.\\
Okinawa, Japan\\
\texttt{dongqi.han@oist.jp}
\And
\bf{Kenji Doya}\\
Neural Computation Unit\\
Okinawa Inst. of Sci. and Tech.\\
Okinawa, Japan\\
\texttt{doya@oist.jp}
}

\begin{document}

\maketitle

\begin{abstract}
In real-world applications of reinforcement learning (RL), noise from inherent stochasticity of environments is inevitable. However, current policy evaluation algorithms, which plays a key role in many RL algorithms, are either prone to noise or inefficient. To solve this issue, we introduce a novel policy evaluation algorithm, which we call Gap-increasing RetrAce Policy Evaluation (GRAPE). It leverages two recent ideas: (1) gap-increasing value update operators in advantage learning for noise-tolerance and (2) off-policy eligibility trace in Retrace algorithm for efficient learning. We provide detailed theoretical analysis of the new algorithm that shows its efficiency and noise-tolerance inherited from Retrace and advantage learning. Furthermore, our analysis shows that GRAPE's learning is significantly efficient than that of a simple learning-rate-based approach while keeping the same level of noise-tolerance. We applied GRAPE to control problems and obtained experimental results supporting our theoretical analysis.
\end{abstract}

% With the second editing by Dr. Aird
%In real-world applications of reinforcement learning (RL), noise from inherent stochasticity of environments is inevitable. However, current policy evaluation algorithms, which serve essential functions in many RL algorithms, are either prone to noise or are inefficient. To solve this problem, we introduce a novel policy evaluation algorithm that  we call Gap-increasing RetrAce Policy Evaluation (GRAPE). It leverages two recent ideas: (1) gap-increasing value update operators in advantage learning for noise-tolerance and (2) off-policy eligibility tracing in Retrace algorithms for efficient learning. We provide detailed theoretical analysis of the new algorithm to show its efficiency and noise-tolerance. Furthermore, our analysis shows that GRAPE's learning is significantly more efficient than that of a simple learning-rate-based approach, while maintaining the same level of noise-tolerance. We applied GRAPE to control problems and obtained experimental results supporting our theoretical analysis.

\section{INTRODUCTION}
Policy evaluation is a key problem in Reinforcement Learning (RL) because many algorithms require a value function for policy improvement \citep{sutton_introRL_2018}. For example, some popular deep RL algorithms are based on actor-critic algorithms, which require a value function \citep{lillicrap_ddpg_2016,wang_acer_2016,mnih_a3c_2016}. However, current policy evaluation algorithms are unsatisfactory since They are either inefficient or prone to noise originating from stochastic rewards and state transitions.

For example, a multi-stage lookahead algorithm called Retrace is efficient in that it is off-policy, uses low-variance updates thanks to truncated importance sampling ratios, and allows control of bias-variance trade-off \citep{munos_retrace_2016}. Retrace achieved state-of-the-art performance on different kinds of RL tasks \citep{wang_acer_2016}. However, Retrace is prone to noise, as shown in Section~\ref{sec:retrace proneness}. Thus, the use of a higher $\lambda$, which results in larger variance of updates, leads to poor performance.

While policy evaluation versions of Dynamic Policy Programming (DPP) \citep{azar_dpp_2012,rawlik_thesis_2013} and Advantage Learning (AL) \citep{baird_thesis_1999,bellemare_action_gap_2016} are noise-tolerant, they do not allow control of bias-variance trade-off because they are single-stage lookahead algorithms.

A simple approach to handle noise is to use a partial update by a learning rate (see \citep{sutton_introRL_2018} for experimental results). We call such an approach learning-rate-based (LR-based). As we argue in Section~\ref{sec:why not learning rate}, although the LR-based approach is noise-tolerant, it suffers from unsatisfactorily slow learning.

To maintain both noise-tolerance and learning efficiency, we propose a new policy evaluation algorithm, called Gap-increasing RetrAce Policy Evaluation (GRAPE), combining Retrace and AL. Theoretical analysis shows that GRAPE is noise-tolerant without significantly sacrificing learning speed and efficiency of Retrace. The theoretical analysis also includes a comparison of GRAPE to Retrace with a learning rate, which emphasizes GRAPE's capacity to learn faster than Retrace with a learning rate. Finally, we demonstrate experimentally that our algorithm outperforms Retrace in noisy environments. These theoretical and experimental results suggest that our algorithm is a promising alternative to previous algorithms.

\section{PRELIMINARIES}
\label{sec:notations and definitions}
We consider finite state-action infinite-horizon Markov Decision Processes (MDPs) \citep{sutton_introRL_2018} defined by a tuple $(\set{X}, \set{A}, P_0, P, \gamma)$, where $\set{X}$ and $\set{A}$ are the finite state and action space,\footnote{Our theoretical results can be extended to a case where $\set{X}$ and $\set{A}$ are compact subsets of finite dimensional Euclid spaces.}, $P_0: \set{X} \rightarrow [0, 1]$ is an initial state distribution, $P: \set{X} \times [-r_{max}, r_{max}] \times \set{X} \times \set{A} \rightarrow [0, 1]$ is the state transition probability with $r_{max} \in (0,\infty)$, and $\gamma \in [0, 1]$ is the discount factor. Their semantics are as follows: at time step $t$, an agent executes an action $a_t \sim \pi (\cdot | x_t)$, where $\pi$ is a policy, and $x_t$ is a state at time $t$. Then, state transition to $x_{t+1}$ occurs with a reward $r_t$ such that $x_{t+1}, r_{t+1} \sim P(\cdot, \cdot | x_t, a_t)$. This process continues until an episode terminates (i.e., state transition to a terminal state occurs). When an episode terminates, the agent starts again from a new initial state $x_0' \sim P_0$.

The following functions are fundamental in RL theory: $Q^{\pi} (x, a) := \E^\pi [ \sum_{t \geq 0} \gamma^t r_t \vert x_0=x, a_0=a ]$ and $V^\pi(x) := \E^\pi [ Q^\pi (x, a) \vert x_0=x ]$, where the superscript $\pi$ on $\E$ indicates that $a_t \sim \pi(\cdot|x_t)$. The former $Q^\pi$ and latter $V^{\pi}$ are called the state-value and Q-value functions for a policy $\pi$, respectively. The aim of the agent is to find an optimal policy $\pi^*$ that satisfies $V^{\pi*} (x) := V^* (x) \geq V^\pi(x)$ for any policy $\pi$ and state $x$. The Q-value and advantage function $A^{\pi} (x, a) := Q^{\pi} (x, a) - V^{\pi} (x)$ play a key role in policy improvement in various RL algorithms. We let $r (x, a)$ denote an expected immediate reward function $\E [ r_0 | x_0=x, a_0=a]$, which is assumed to be bounded by $r_{max}$. Note that $V^{\pi}$ and $Q^{\pi}$ are bounded by $V_{max} := r_{max} / (1 - \gamma)$. We let $\set{Q}$ and $\set{V}$ denote bounded functions over $\set{X} \times \set{A}$ and $\set{X}$, respectively. $\set{Q}$ and $\set{V}$ can be understood as vector spaces over a field $\R$. A sum of $Q \in \set{Q}$ and $V \in \set{V}$ is defined as $(Q + V)(x, a) := Q(x, a) + V(x)$. In this paper, we measure distance between functions $f$ and $g$ by $l_\infty$-norm $\| f - g \| := \max_{s \in \set{S}} |f (s) - g (s)|$, where $\set{S}$ is a domain of $f$ and $g$. An operator $\op{O}$ from a functional space $\set{F}$ to $\set{F}$ is a contraction with modulus $L \in [0, 1)$ around a fixed point $f^*$ if $\| \op{O} f - f^* \| \leq L \| f - f^* \|$ holds for any function $f \in \set{F}$.

\subsection{APPROXIMATE DYNAMIC PROGRAMMING}\label{subsec:dynamic programming}
In this paper, we consider the following setting frequently used in off-policy RL: we have an experience buffer $\set{D}$ to which a tuple $(x, a, r, y, d)$ - a state, action, reward, subsequent state, and binary value with $1$ indicating that $y$ is a terminal state, is constantly appended, as an agent gets new experience until $\set{D}$ is full. When $\set{D}$ is full, the oldest tuple at the beginning is removed, and a new one is appended to the end. It returns samples when queried, and values and/or policy updates are carried out using samples. (How samples are obtained depends on algorithms to be used.)

The buffer is understood as a device that returns samples of tuples $(x, a, r, y, d)$. One of the simplest policy evaluation algorithms under this setting is shown in Algorithm~\ref{algorithm:phased TD0}. It approximates a dynamic programming (DP) algorithm that computes $Q^\pi$ by recursively updating a function $Q_k \in \set{Q}$ according to $Q_{k+1} := \op{T^\pi} Q_k$, where the update is pointwise, and $\op{T^\pi}: \set{Q} \rightarrow \set{Q}$ is the Bellman operator $\pi$ defined such that $\left( \op{T^\pi} Q \right) (x, a) := r(x, a) + \gamma \E_{x_1, a_1 \sim \pi} [Q(x_1, a_1) | x_0=x, a_0=a]$. We call this DP algorithm and Algorithm~\ref{algorithm:phased TD0} exact and approximate phased TD($0$), respectively \citep{kearns_TDlambda_bias_variance_2000}. However, we frequently omit their qualifiers "exact" and "approximate" for brevity.

\begin{algorithm}[t]                    
	\caption{Phased TD($0$)}         
	\label{algorithm:phased TD0}                          
	\begin{algorithmic}                  
		\REQUIRE A buffer $\set{D}$, function class $\set{F}$, policy $\pi$.
		\STATE Initialize $Q_0$.
		\FOR{$k = 0, 1, \ldots, K$}
		\FOR{$t = 0, 1, \ldots$ until a terminal state is reached}
		\STATE Observe a state $x_t$.
		\STATE Take an action $a_t \sim \pi (\cdot | x_t)$.
		\STATE Get and observe a reward and next state $r_t, x_{t+1}$.
		\STATE $d_t \gets 1$ if $x_{t+1}$ is terminal otherwise $d_t \gets 0$.
		\STATE Append $(x_t, a_t, r_t, x_{t+1}, d_t)$ to $\set{D}$.
		\ENDFOR
		\STATE Sample $(x_i, a_i, r_i, y_i, d_i) \sim \set{D}$.
		\STATE $Q_{k+1} \gets \argmin_{Q \in \set{F}} \sum_{i=1}^{N} \left| \Delta_i \right|^2$, where $\Delta_i$ is $r_i + \gamma (1 - d_i) \sum_{b \in \set{A}} \pi (a|y_i) Q_k (y_i, b) - Q (x_i, a_i)$.
		\STATE Discard samples in $\set{D}$.
		\ENDFOR
	\end{algorithmic}
\end{algorithm}

Approximate phased TD($0$) is an approximation in a sense that $\op{T^{\pi}} Q_k$ is estimated by samples, and a function approximator is used for $Q_{k+1}$. \textit{In this paper, we refer to errors in updates caused by finite-sample estimation of $\op{T^{\pi}} Q_k$ as noise}. Such errors stem from stochasticity of the environment in case of model-free RL.

In theoretical analysis, we use error functions that abstractly express update errors. In the current example, phased TD($0$)'s non-exact update rule is given as $Q_{k+1} := \op{T^\pi} Q_k + \varepsilon_k$, where $\varepsilon_k \in \set{Q}$ is the error function at $k$-th iteration. Analysis of how $\varepsilon_k$ at each iteration $k \in \{0, 1, \ldots, K \}$ affects final performance (in our case, measured by \eqref{eq:aim of theoretical analysis}) is called error propagation analysis and a typical way to analyze approximate DP algorithms \citep{munos05,munos07,FarahmandThesis,scherrer2012_nips_on_the_use_of_nonstationary,azar_dpp_2012}.

\subsection{RETRACE: OFF-POLICY MULTI-STAGE LOOKAHEAD POLICY EVALUATION}\label{subsec:previous algorithms}
In addition to phased TD($0$), many policy evaluation DP algorithms have been proposed \citep{sutton_introRL_2018}. Munos et al.~provided a unified view of those algorithms \citep{munos_retrace_2016}. Suppose a target policy $\pi$, the Q-value function of which we want to estimate, and behavior policy $\mu$, with which data are collected. Let $\rho (x, a)$ denote the importance sampling ratio $\pi (a|x) / \mu (a|x)$, which is assumed to be well-defined. We define an operator $\op{P^{c_0 \mu}}: \set{Q} \rightarrow \set{Q}$ such that $( \op{P^{c_0 \mu}} Q ) (x, a) := \E^\mu [ c (x_1, a_1) Q(x_1, a_1) \vert x_0=x, a_0=a ]$, where $c_0: \set{X} \times \set{A} \rightarrow \left[0, \rho (x, a) \right]$ corrects the difference between $\pi$ and $\mu$. Munos et al. showed that an operator $\op{R_{\lambda}^{c_0 \mu}}$ in the following equation is a contraction around $Q^{\pi}$ with modulus $\gamma$; thus, $Q_k$ obtained by the following rule uniformly converges to $Q^\pi$:
\begin{align}\label{eq:retrace update}
	Q_{k+1}
	:= \op{R_{\lambda}^{c_0 \mu}} Q_k,
\end{align}
where $\lambda \in [0, 1]$, and $\op{R_{\lambda}^{c_0 \mu}}: \set{Q} \rightarrow \set{Q}$ is an operator such that $\op{R_{\lambda}^{c_0 \mu}} Q := Q + \sum_{t=0}^\infty \lambda^t \gamma^t \left( \op{P^{c \mu}} \right)^t \left( \op{T^{\pi}} Q - Q \right)$. Approximate Retrace can be implemented similarly to Algorithm~\ref{algorithm:phased TD0}, but trajectories must be sampled from $\set{D}$.

Depending on $c_0$, various algorithms are reconstructed. For example, tree-backup is obtained when $c(x, a) = \pi (a|x)$, while phased TD($\lambda$) with importance sampling is obtained when $c_0 (x, a) = \rho(x, a)$ \citep{precup_tb_2000}. In particular, Munos et al. proposed to use $c_0 (x, a) = \min \{1, \rho(x, a) \}$ and called the resultant algorithm Retrace. When we mean this choice of $c_0$, we use $c$ to differentiate from other choices, and thus, $\op{R_{\lambda}^{c_0 \mu}}$ is denoted as $\op{R_{\lambda}^{c \mu}}$ instead. 

\begin{remark}
	The following generalization of Retrace update works too as $\left\| Q^{\pi} - Q_{k+1} \right\| \leq \gamma \left\| Q^{\pi} - Q_k \right\|$ holds: $Q_{k+1} := \sum_{i=1}^{I} p_i \op{R_{\lambda}^{c_0 \mu_i}} Q_k$, where $\mu_i$ is $i$-th behavior policy, and $p_i \in [0, 1], \sum_i p_i = 1$. It is suitable to combination with a buffer $\set{D}$ containing trajectories obtained by following several policies.
\end{remark}

\section{RETRACE'S PRONENESS TO NOISE}\label{sec:retrace proneness}
% \begin{figure}[t]
% 	\begin{minipage}[b]{0.3\columnwidth}
% 		\includegraphics[width=\textwidth]{images/frozenlake.png}
% 	\end{minipage}
% 	\begin{minipage}[b]{0.68\columnwidth}
% 		\caption{$8\times8$ FrozenLake. Blue grids are slippery but safe states, while black grids are terminal states with no rewards. An agent obtains a reward $1$ when it reaches a goal, G, (bottom right) from a start state S (top left).}\label{fig:frozenlake}
% 	\end{minipage}
% \end{figure}

\begin{figure}[h]
    \centering
    \includegraphics[width=0.5\columnwidth]{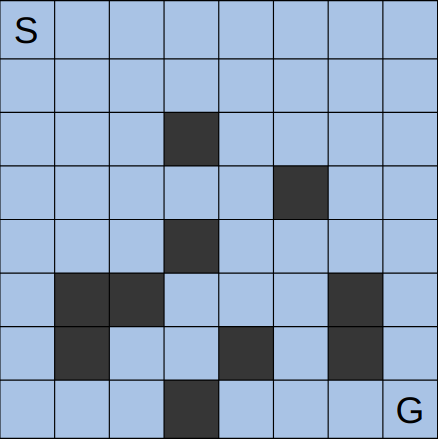}
    \caption{$8\times8$ FrozenLake. Blue grids are slippery but safe states, while black grids are terminal states with no rewards. An agent obtains a reward $1$ when it reaches a goal, G, (bottom right) from a start state S (top left).}
    \label{fig:frozenlake}
\end{figure}

In \citep{munos_retrace_2016}, the convergence of exact Retrace is proven. However, in a simple experiment with a lookup table in $8 \times 8$ FrozenLake in OpenAI Gym \citep{openai_gym} shown in Fig.~\ref{fig:frozenlake}, we found Retrace's proneness to noise.

The experiment is done as follows: first, $\mu$ and $\pi$ are sampled from a Dirichlet distribution with concentration parameters all set to $1$. Then, using the policies, matrices $\op{P^{c\mu}}$ and $\op{P^{\pi}}$ are constructed. Using $\op{P^{c\mu}}$, $\op{P^{\pi}}$ and an expected reward function $r$, $(Q_{k+1})(x, a)$ is computed as a sum of $(\op{R_{\lambda}^{c \mu}} Q_k)(x, a)$ and Gaussian noise $\varepsilon_k (x, a) \sim N(0, \sigma)$, where $\lambda = 0.8$, and $\gamma = 0.99$. Similar results are obtained regardless of values of $\lambda$ and $\gamma$. The standard deviation $\sigma \in \{0.0, 0.4, 0.8\}$ is varied to investigate the effect of noise intensity. An initial function is $Q_0 (x, a) \sim N(0, 1)$.

To measure the performance of Retrace, we used normalized root mean squared error (NRMSE). Let $e_K$ be
\begin{equation*}
	e_K := \frac{1}{\left| \set{X} \times \set{A} \right|} \sum_{(x, a) \in \set{X} \times \set{A}} \left( A^{\pi} (x, a) - A_K (x, a) \right)^2,
\end{equation*}
where $A_K (x, a) := Q_K (x, a) - \sum_{a \in \set{A}} \pi (a|x) Q_K (x, a)$. NRMSE is defined by $e_K / e_0$.

\begin{figure}[t]
	\includegraphics[width=\columnwidth]{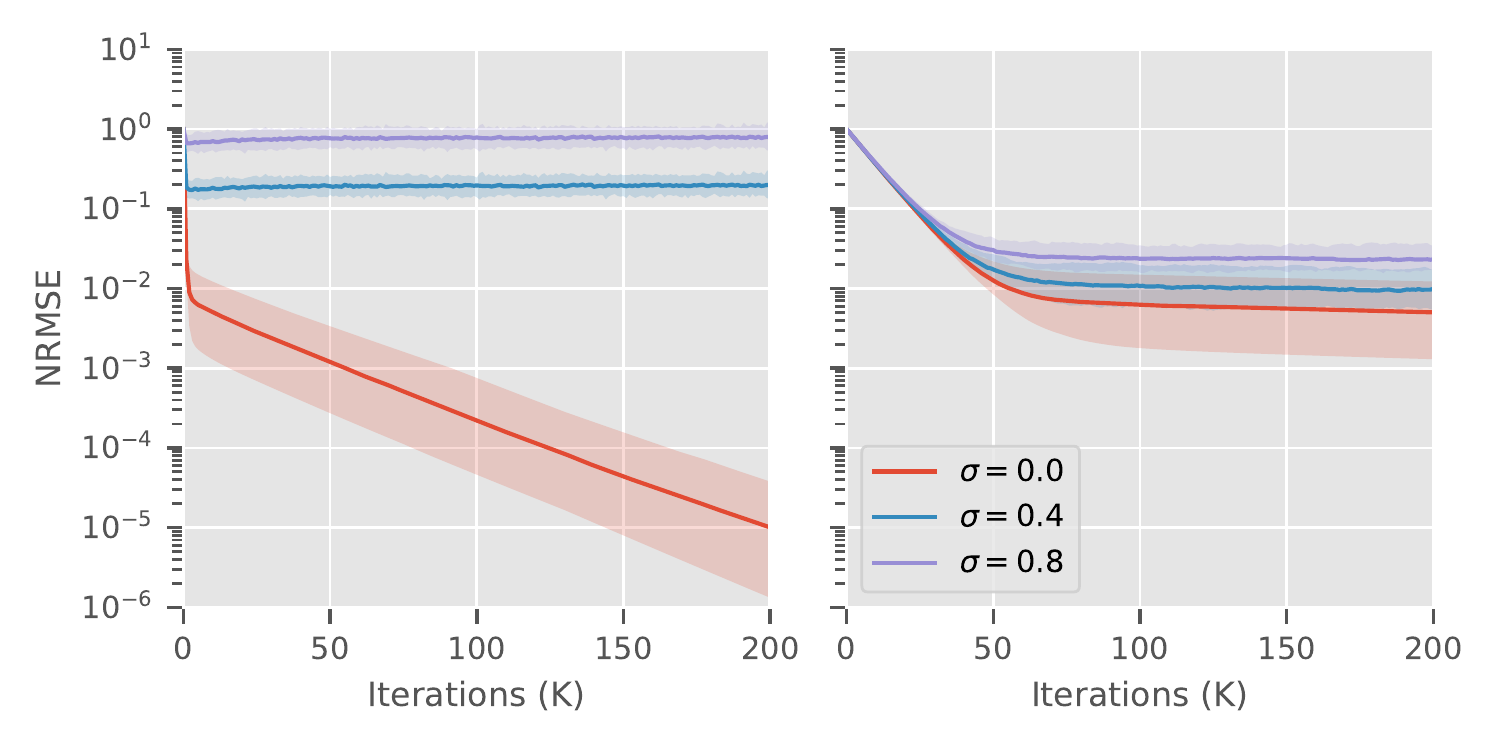}
	\caption{Experimental results in $8\times8$ FrozenLake with Retrace (left panel) and Retrace with a learning rate (right panel) using DP updates. Lines indicate the median of NRMSE (lower is better) over $100$ experiments, and the shaded area shows the $95$ percentile. Colors indicate noise intensity $\sigma$. Note that the vertical axis is in log-scale.}\label{fig:retrace_comparison}
\end{figure}

The left panel of Fig.~\ref{fig:retrace_comparison} visualizes performance of Retrace measured by this NRMSE with varying noise intensity. The result shows that Retrace suffers from noise. In particular, when $\sigma=0.8$, NRMSE is approximately $1$, meaning almost no learning occurs.

\section{SLOW LEARNING DUE TO LEARNING RATES}\label{sec:why not learning rate}
As we have now seen, the Retrace algorithm is prone to noise. A simple approach to handle noise is to use a learning-rate. We call such an approach learning rate (LR)-based. For example, the update rule of phased TD($0$) with a learning rate is
\begin{align}\label{eq:example of lr based approach}
	Q_{k+1} := \eta_k \odot \op{T^{\pi}} Q_k + \left( 1 - \eta_k \right) \odot Q_k,
\end{align}
where $\eta_k: \set{X} \times \set{A} \rightarrow [0, 1]$ is a learning rate, and $\odot$ is element-wise multiplication, i.e., $(\left( 1 - \eta_k \right) \odot Q_k)(x, a) = \left( 1 - \eta_k (x, a) \right) Q_k(x, a)$. This generalized notion of a learning rate is frequently used in theoretical analysis \citep{bertsekas_ndp_1996,singh_2000,even_dar_q_learning_2004}. The LR-based approach includes various algorithms. For example, the standard online TD($0$) is obtained when $k = t$, and $\eta_k (x, a) \neq 0$ if and only if $x$ and $a$ is visited at time $t$.

%LR-based approach attains, as expected, noise-tolerance. For simplicity, let us assume that $Q_0 (x, a) = 0$ and $\eta_k (x, a) = \eta \in (0, 1]$ for all $k$ and state-action pairs $(x, a) \in \set{X} \times \set{A}$. Let us suppose that due to noise, the update \eqref{eq:example of lr based approach} becomes
%\begin{align}\label{eq:approximate TD update}
%	\vec{Q_{k+1}}
%	&:= \eta \left( \op{T^{\pi}} \vec{Q_k} + \vec{\varepsilon_k} \right) + \left( 1 - \eta \right) \vec{Q_k} \nonumber\\
%	&= \eta \sum_{l=0}^{k} (1 - \eta)^l \left( \op{T^{\pi}} \vec{Q_{k-l}} + \vec{\varepsilon_{k-l}} \right).
%\end{align}
%Because $\vec{Q^\pi} = \eta \sum_{l=0}^{K} (1 - \eta)^l \vec{Q^{\pi}} + (1 - \eta)^{K+1} \vec{Q^{\pi}}$,
%\begin{align*}
%	\left\| \vec{Q^\pi} - \vec{Q_{K+1}} \right\|
%	&\leq \eta \gamma \sum_{k=0}^{K} (1 - \eta)^k \left\| \vec{Q^{\pi}} - \vec{Q_{K-k}} \right\|\\
%	&\hspace{3em}+ \left\| E_K \right\| + (1 - \eta)^{K+1} V_{max},
%\end{align*}
%where $E_L := \eta \sum_{l=0}^{L} (1 - \eta)^l \vec{\varepsilon_{L-l}}$. By induction, it is easy to show that an upper bound of the last line consists of many $\eta \left\| E_L' \right\|$, where $E_L' := \sum_{l=0}^{L} (1 - \eta)^l \vec{\varepsilon_{L-l}}$. As $\eta E_L' (s, a)$ is a moving average of noise $\varepsilon_{L-l} (s, a)$, it is expected that $\eta E_L' \approx 0$. Hence, LR-based approach attains noise-tolerance.

Although the LR-based approach is noise-tolerant, it often demands more iterations and thus leads to slow learning. For simplicity, let us assume that $Q_0 (x, a) = 0$ and $\eta_k (x, a) = \eta$ for any state $x$ and action $a$. Then, the update \eqref{eq:example of lr based approach} becomes
\begin{align}\label{eq:TD update with learning rate}
	Q_{k+1}
	:= \eta \op{T^{\pi}} Q_k + \left( 1 - \eta \right) Q_k  = \eta \sum_{l=0}^{k} \op{\Gamma}^l r,
\end{align}
where $\op{\Gamma}: \set{Q} \rightarrow \set{Q}$ is defined as $\op{\Gamma} Q := \left( 1 - \eta \right) Q + \eta \gamma \op{P^{\pi}} Q$. As $Q^\pi = \eta \sum_{k=0}^{K-1} \op{\Gamma}^k r + \op{\Gamma}^{K} Q^{\pi}$,
\begin{align}\label{eq:convergence rate of TD 0}
	\left\| Q^{\pi} - Q_{K} \right\| \leq \left( 1 - \eta (1 - \gamma) \right)^{K} V_{max}.
\end{align}
As this upper bound holds with equality when $\op{P^\pi}$ is an identity operator $\op{I}$, it is not improvable. (For example, $\op{P^\pi} = \op{I}$ when an environment has only one state and action). Therefore, the convergence rate is $O( ( 1 - \eta (1 - \gamma) )^K )$. Considering that $\gamma \approx 1$ and $\eta \approx 0$ in many cases, $1 - \eta (1 - \gamma)$ is close to $1$. Thus, the LR-based approach is noise-tolerant at the sacrifice of learning efficiency.

To confirm this argument, we conducted experiments using Retrace with a learning rate. The right panel of Fig.~\ref{fig:retrace_comparison} shows the results. It illustrates the tolerance of the LR-based approach to noise as well as its slow learning. (The red line seems to be flat, but it has a very slight slope, indicating the unsatisfactorily slow learning of the LR-based approach.)

\section{GAP-INCREASING RETRACE ADVANTAGE POLICY EVALUATION (GRAPE)}
Section~\ref{sec:why not learning rate} discussed noise-tolerance of the LR-based approach at the sacrifice of learning efficiency. Is it possible to tame noise while maintaining efficiency? In this section, we affirmatively answer the question with a gap-increasing policy evaluation algorithm, called GRAPE, inspired by AL and DPP \citep{baird_thesis_1999,azar_dpp_2012,rawlik_thesis_2013,bellemare_action_gap_2016}.

Suppose target and behavior policies $\pi, \mu$ and two real numbers $\alpha, \lambda \in [0, 1]$. Let $\op{G^{c \mu}_\lambda}$ denote an operator $\op{G^{c \mu}_\lambda} Q := \op{T^{\pi}} Q + \sum_{t=0}^\infty \gamma^{t+1} \lambda^{t+1} ( \op{P^{c \mu}} )^t \op{P^{\pi}} ( \op{T^{\pi}} Q - Q )$. GRAPE's update rule is the following: suppose an initial function $\Psi_0 \in \set{Q}$. $\Psi_1 \in \set{Q}$ is defined as $\op{G^{c \mu}_\lambda} \Psi_0$. $\Psi_{k+1} \in \set{Q}, k \in \{1, 2, \ldots\}$ is recursively defined by
\begin{align}\label{eq:grape update}
	\Psi_{k+1} := \op{G^{c \mu}_\lambda} \Psi_k + \alpha \Phi_k,
\end{align}
where $\Phi_k (x, a) := \Psi_k (x, a) - \sum_{a \in \set{A}} \pi (a|x) \Psi_k (x, a)$

This update rule is very similar to that of AL except for the use of $\op{G^{c \mu}_\lambda}$ rather than $\op{T^\pi}$. The reason why we use $\op{G^{c \mu}_\lambda}$ instead of $\op{R^{c \mu}_\lambda}$ is that this choice of the variant combined with our proof technique allows us to theoretically show GRAPE's noise-tolerance later in the theoretical analysis.

\begin{remark}
	The reason for a different update $\Psi_1 = \op{G^{c \mu}_\lambda} \Psi_0$ is to make the argument of Remark~\ref{remark:on reuse of Psi} valid. Yet the convergence rate and noise-tolerance remain the same, even if the update~\eqref{eq:grape update} is used for $k=0$ too.
\end{remark}

For model-free GRAPE, several variants can be conceived, depending on how to estimate $\left( \op{G^{c \mu}_\lambda} \Psi_k \right)(x, a)$ with samples. Appendix~\ref{sec:discussion on which estimator to be used} provides a brief discussion, based on which, we propose the following estimator:
\begin{align}\label{eq:model-free estimator}
	&r_0 + \gamma \left( \op{\pi} \Psi \right) (x_1)\\
	&\hspace{1em}+ \sum_{t=0}^\infty \gamma^{t+1} \lambda^{t+1} \prod_{u=1}^{t} c (x_u, a_u) \rho (x_{t+1}, a_{t+1}) \Delta_{t+1},\nonumber
\end{align}
where we omit an iteration index $k$ of $\Psi_k$ to avoid notational confusion with a time index $t$, $x_0=x$, $a_0=a$, actions are selected according to $\mu$, $\prod_{u=1}^{0} c (x_u, a_u) = 1$, and $\Delta_t$ is defined as
\begin{align*}
	 r_t + \gamma \left( \op{{\pi}} \Psi \right) (x_{t+1}) - (1 - \alpha) \Psi (x_t, a_t) - \alpha \left( \op{\pi} \Psi \right) (x_t).
\end{align*}
This $\Delta_t$ is an unbiased estimate of $(\op{T} \Psi) (s, a) - \Psi(s, a)$ at time step $t$. However, as shown in Theorem~\ref{theorem:convergence of Psi_k} later, $\lim_{K \rightarrow \infty} \Psi_k = V^\pi + A^\pi / (1 - \alpha)$. Therefore, depending on $\alpha$, $\Psi_k$ may have large values. Accordingly, vanilla estimator $r_t + \gamma \left( \op{{\pi}} \Psi \right) (x_{t+1}) - \Psi (x_t, a_t)$ may have too large a variance. In contrast, $\Delta_t$ may not.

Algorithm~\ref{algo:grape pseudo-algorithm} is a model-free implementation of GRAPE. Note that this algorithm is an approximation of GRAPE since sweeping a whole buffer $\set{D}$ to exactly compute update targets is costly.

\begin{algorithm}[t]                    
	\caption{GRAPE}
	\label{algo:grape pseudo-algorithm}                       
	\begin{algorithmic}
		\REQUIRE  Contiguous samples $(x_t, a_t, r_t, x_{t+1}, \mu_t, d_t)$ from a buffer $\set{D}$, where $t \in \{0, 1, \ldots, T\}$, a current value function $\Psi$, and a target policy $\pi$.
		\STATE $b_{T+1} \gets 0$.
		\FOR{\text{$t$ from $T$ to $0$}}
		    \STATE $\rho_t \gets \pi(a_t | s_t) / \mu_t$, $c_t \gets \min \{1, \rho_t \}$.
		    \STATE $\reallywidehat{\op{T^\pi} \Psi}_t \gets r_t + \gamma (1 - d_t) \sum_{b \in \set{A}} \Psi (x_{t+1}, b)$.
		    \STATE $\Phi_t \gets \Psi (x_t, a_t) - \sum_{b \in \set{A}} \pi (b|x) \Psi (x_t, b)$.
		    \STATE $\reallywidehat{\op{G^{c \mu}_\lambda} \Psi}_t \gets \reallywidehat{\op{T^\pi} \Psi}_t + \alpha \Phi_t + \gamma \lambda b_{t+1}$.
		    \STATE $b_t \gets \rho_t \left( \reallywidehat{\op{T^\pi} \Psi}_t - \Psi (s_t, a_t) + \alpha \Phi_t \right) + \gamma \lambda c_{t} b_{t+1}$ if $d_t = 0$ otherwise $b_t \gets 0$.
		\ENDFOR
		\RETURN Update targets $(\reallywidehat{\op{G^{c \mu}_\lambda} \Psi}_0, \ldots, \reallywidehat{\op{G^{c \mu}_\lambda} \Psi}_T)$.
	\end{algorithmic}
\end{algorithm}

%Although we could theoretically show only GRAPE's noise-tolerance, a similar algorithm with the following update rule empirically worked well too:
%\begin{align}\label{eq:rgrape update}
%	\Psi_{k+1} := \Psi_k + \sum_{t=0}^{\infty} \gamma^t \lambda^t \left( \op{P^{c \mu}} \right)^t \left( \op{T^{\pi}} \Psi_{k} - \Psi_{k} + \alpha \Phi \right).
%\end{align}
%We call this algorithm RGRAPE. It is not difficult to confirm that it has a unique fixed point $V^{\pi} (x, a) + A^{\pi} (x, a) / (1 - \alpha)$, to which GRAPE converges when $\alpha \neq 1$.

\subsection{THEORETICAL ANALYSIS OF GRAPE}\label{subsec:theoretical analysis of GRAPE}
We theoretically analyze GRAPE to understand its learning behavior. All proofs are deferred to appendices. For simplicity, we assume that $\delta \neq \alpha$, where $\delta$ is defined ine Lemma~\ref{lemma:G is a contraction}.

The following lemma shows that $\op{G^{c \mu}_\lambda}$ is a contraction, and our theoretical analysis relies heavily upon it.
\begin{lemma}\label{lemma:G is a contraction}
	$\op{G^{c \mu}_\lambda}$ is a contraction around $Q^\pi$ with modulus $\delta := \gamma \left(1 - \lambda \left(1 - \gamma\right) \right)$.
\end{lemma}

\begin{remark}
	As argued in Remark~1 of \citep{munos_retrace_2016}, a modulus ($\delta$ in our case) of Retrace is smaller when $\pi$ and $\mu$ are close. Similarly, $\delta$ is smaller when $\pi$ and $\mu$ are close. In other words, $\gamma \left( 1 - \lambda \left(1 - \gamma \right) \right)$ is the worst-case modulus.
\end{remark}

\subsubsection{Convergence}
We have the following result regarding exact GRAPE.
\begin{theorem}\label{theorem:convergence of Psi_k}
	GRAPE has the following convergence property:
	\begin{gather*}
		\lim_{K \rightarrow \infty} \Phi_K / A_K = A^{\pi},\\
		\lim_{K \rightarrow \infty} \Psi_K / A_K = A^{\pi} + (1 - \alpha) V^{\pi},
	\end{gather*}
	where $A_K := \sum_{k=0}^K \alpha^k$. Moreover, their convergence rates are $O(\max\{\alpha, \delta\}^K)$ when $\alpha \neq 1$ and $O(K^{-1})$ when $\alpha = 1$.
\end{theorem}

\begin{figure}[t]
	\centering
	\includegraphics[width=\columnwidth]{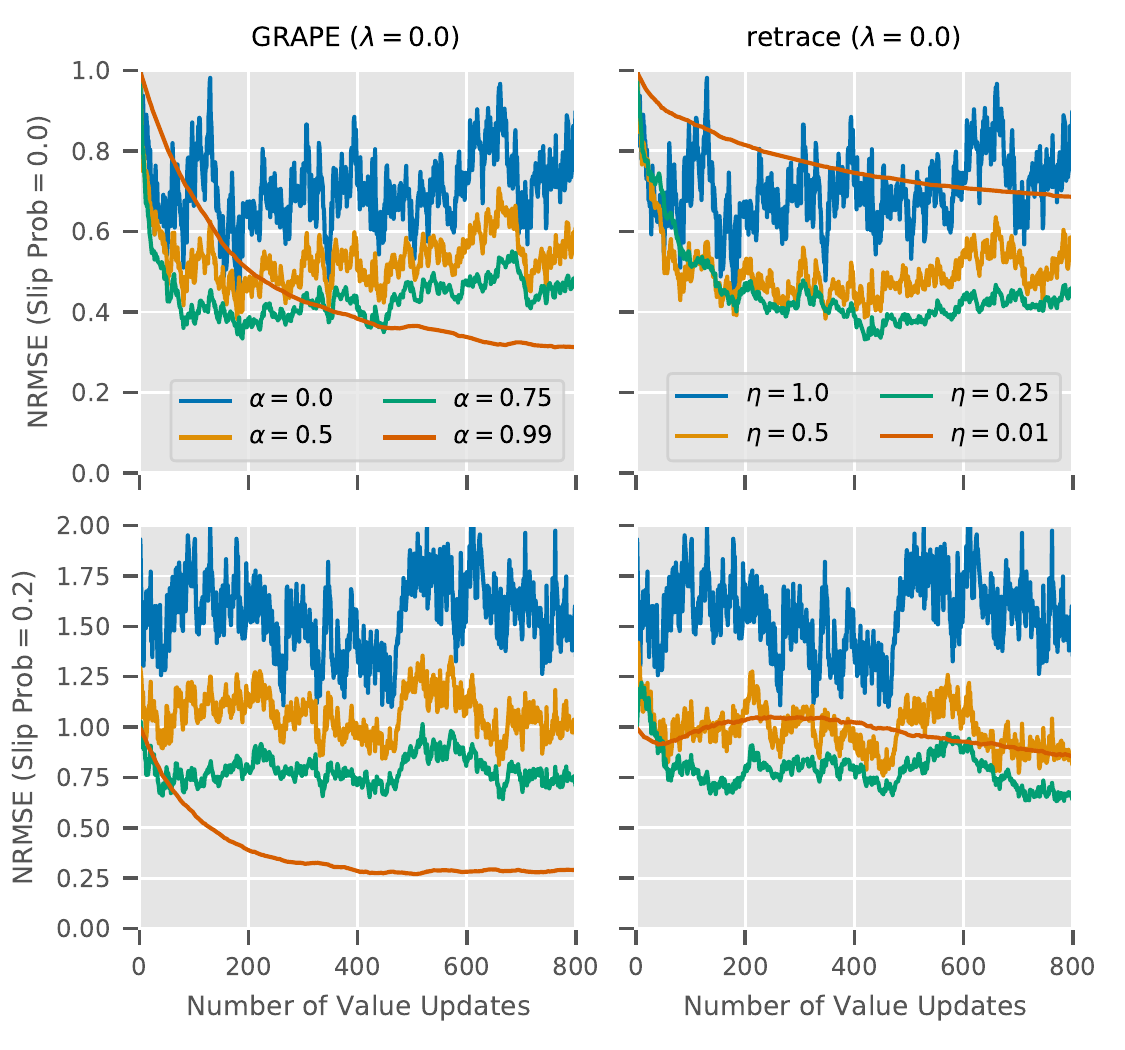}
	\caption{Policy evaluation performance comparison of GRAPE and Retrace with a learning rate in NChain. The horizontal axes show the number of value updates. The vertical axes show NRMSE (lower is better). Lines show mean performance over twenty-four trials. For visibility, we omit error bars. The first row shows results of GRAPE and Retrace when slip prob is $0.0$. $\alpha$ and $\eta$ are indicated by the legends. $\lambda$ is fixed to $0.0$. The second row is the same except that slip prob is increased to $0.2$.}
	\label{fig:performance comparison in nchain}
\end{figure}

As we show experimentally later (Fig.~\ref{fig:performance comparison in nchain}), noise-tolerance of GRAPE is approximately same as that of the LR-based approach when $\alpha = 1 - \eta$. Thus, we can compare the convergence rate of GRAPE and Retrace with a learning rate by comparing $1-\eta (1 - \delta)$ and $\max\{\alpha, \delta\}$ in which $\alpha = 1-\eta$. Suppose that $\alpha =  1-\eta < \delta$. Then, GRAPE's convergence rate is $O(\delta^K)$, whereas that of the LR-based approach is $K$-th power of $1-\eta (1 - \delta) = \delta + (1-\eta) (1 - \delta) = \delta + \alpha (1 - \delta) > \delta$. Accordingly, GRAPE learns faster than the LR-based approach does. Particularly, in this example, GRAPE's faster learning is eminent when $\alpha \approx 1$.

Interestingly, while a fixed point of previous policy evaluation algorithms is $Q^\pi$, GRAPE's fixed point is $V^\pi (x) + A^\pi (x, a) / ( 1 - \alpha)$ when $\alpha \neq 1$. Thus, in GRAPE, $A^\pi$ is enhanced by a factor of $1 / ( 1 - \alpha )$. This is the reason why we call GRAPE \textit{gap-increasing} Retrace; Q-value differences, or action-gaps, are increased. In case of AL, its fixed point is $V^* (x) + A^* (x, a) / ( 1 - \alpha )$, which is indicative of the point to which GRAPE converges \citep{kozuno_cvi_2017}.

This gap-increasing property might be beneficial when RL is applied to a system operating at a fine time scale, as argued in \citep{baird_thesis_1999,bellemare_action_gap_2016}. Briefly, in such a situation, changes of states caused by an action at one time step are small. Consequently, so are action-gaps. Hence, a function approximator combined with a previous policy evaluation algorithm mainly approximates $V^\pi$ rather than $A^\pi$ (because it tries to minimize error between $Q^\pi = V^\pi + A^\pi$ and an estimated Q-value function). However, $A^\pi$ is the one truly required to improve a policy.

\subsubsection{Error Propagation Analysis}
A more interesting question on GRAPE is how update errors affect performance. To this end, we consider error functions $\varepsilon_k$ (see Section~\ref{subsec:dynamic programming}) such that non-exact GRAPE updates are given by $\Psi_1 := \op{G^{c \mu}_\lambda} \Psi_0 + \varepsilon_0$ and
\begin{align}\label{eq:agrape update}
	\Psi_{k+1} := \op{G^{c \mu}_\lambda} \Psi_k + \alpha \Phi_k + \varepsilon_k,
\end{align}
where we note that $\varepsilon_k$ may be completely different from $\varepsilon_k$ in Section~\ref{subsec:dynamic programming}; It depends on the algorithm to be used and a function class used for approximating an estimate of $\op{G^{c \mu}_\lambda} \Psi_k + \alpha \Phi_k$. However, when an estimator \eqref{eq:model-free estimator} is used, an order of $\varepsilon_k$ would not be much different from that of previous algorithms.\footnote{Indeed, either $\op{\pi} \Psi_k$ or $(1 - \alpha) \Psi_k + \alpha \op{\pi} \Psi_k$ is used in the estimator. From Lemma~\ref{lemma:simplified Psi_k anoter expression} and \ref{lemma:convergence of q_k} in Appendix~\ref{appendix:proof}, it is easy to deduce that $\lim_{k \rightarrow \infty} \op{\pi} \Psi_k = V^\pi$ in exact GRAPE. On the other hand, Theorem~\ref{theorem:convergence of Psi_k} implies that $\lim_{k \rightarrow \infty} (1 - \alpha) \Psi_k + \alpha \op{\pi} \Psi_k = Q^\pi$. Thus, an order of $\varepsilon_k$ would not be different from that of previous algorithms.}

In this section, we provide an upper bound of
\begin{align}\label{eq:aim of theoretical analysis}
	\left\| A^\pi - \Phi_K / A_K \right\|
\end{align}
expressed by error functions to measure how close $\Phi_K / A_K$ is to $A^\pi$. One may wonder why we do not investigate the distance between $Q^\pi$ and some function $Q$. The reason is that $\left\| Q^\pi - Q \right\|$ might be small even if $Q$ is useless for policy improvement. For example, if $A^\pi$ is small compared to $V^\pi$, then, setting $Q$ to be $V^\pi$ yields a small $\left\| Q^\pi - Q \right\|$.

We have the following theorem that provides an upper bound and implies noise-tolerance of GRAPE.
\begin{theorem}\label{theorem:agrape performance bound}
	GRAPE has the following error bound:
	\begin{align*}
		&\left\| A^\pi - \Phi_K / A_K \right\|\\
		&\hspace{0em} \leq \frac{2 \delta \Gamma_K}{A_K} \left\| V^{\pi} - \op{\pi} \Psi_0 \right\| + 2 \sum_{k=0}^{K-1} \delta^{K-k-1} \left\| E_k / A_K \right\|,
	\end{align*}
	where $\delta$ and $A_K$ are defined in Lemma~\ref{lemma:G is a contraction} and Theorem~\ref{theorem:convergence of Psi_k}, respectively, and we have
	\begin{gather*}
		\Gamma_K := \frac{\alpha^K - \delta^K}{\left(\alpha - \delta \right)}
		\text{, and }
		E_k (x, a) := \sum_{l=0}^{k} \alpha^l \varepsilon_{k-l} (x, a).
	\end{gather*}
\end{theorem}

\begin{remark}
	A generalization of the theorem using other norms is possible. To do so, we need to generalize concentrability coefficients \citep{munos05,munos07,FarahmandThesis,scherrer2012_nips_on_the_use_of_nonstationary}. It is straightforward, but we omit it due to page limitation.
\end{remark}

\begin{remark}\label{remark:on reuse of Psi}
	When a policy evaluation algorithm is combined with a function approximator, it is often the case that $\Psi_K$ is reused after a policy update as an initial function $\Psi_0$. Let us denote a policy before and after the update as $\widetilde{\pi}$ and $\pi$, respectively. Then, it is possible to show (cf. Appendix~\ref{appendix:reuse of Psi}) that
	\begin{align}
		&\left\| V^\pi - \op{\pi} \Psi_0 \right\|\label{eq:reuse of Psi}\\
		&\leq \frac{\sqrt{2} V_{max}}{1 - \gamma} D^{1/2} + \sqrt{2} D^{1/2} \left\| \Psi_0 \right\| + \left\| V^{\widetilde{\pi}} - \op{\widetilde{\pi}} \Psi_0 \right\|,\nonumber
	\end{align}
	where $D := \max_{x \in \set{X}} \E^\pi [ \log ( \pi (a|x) / \widetilde{\pi} (a|x)) ]$ is the maximum Kullback{\textendash}Leibler (KL) divergence. Since in exact GRAPE, $\op{\pi} \Psi_K$ is converging to $V^{\pi}$, the third term is expected to be close to $0$ when the reuse of $\Psi_K$ is done, whereas the first and second terms are close to $0$ when $D$ is small. Therefore, the reuse of $\Psi_K$ as explained above would work well with policy iteration algorithms that try to keep $D$ small. TRPO is a recent popular instance.
\end{remark}

To see GRAPE's noise-tolerance indicated by Theorem~\ref{theorem:agrape performance bound}, suppose that $\varepsilon_k (x, a), k \in \{0, 1, \ldots, K \}$ are i.i.d.~random variables whose mean and variance are $0$ and $1$, respectively. Then, $E_{k-1} (x, a) / A_k$ has a variance $(1 + \alpha^2 + \cdots + \alpha^{2k}) / A_k^2$. It converges to approximately $0.005$ when $\alpha = 0.99$, while it is $1$ when $\alpha=0$. Thus, a higher $\alpha$ leads to a significantly smaller variance. Although $\varepsilon_k (x, a), k \in \{0, 1, \ldots, K \}$ are not i.i.d.~in practice, a similar result is expected to hold in model-free setting, in which updates are estimated from samples.

Note that this argument also shows the ineffectiveness of increasing the number of samples to reduce a variance of $\varepsilon_{i}$. To attain a variance of $\varepsilon_{i}$ as small as $0.005$, around two hundred times more samples are required ($1/0.005 \approx 200$).

\begin{figure}[t]
	\centering
	\includegraphics[width=\columnwidth]{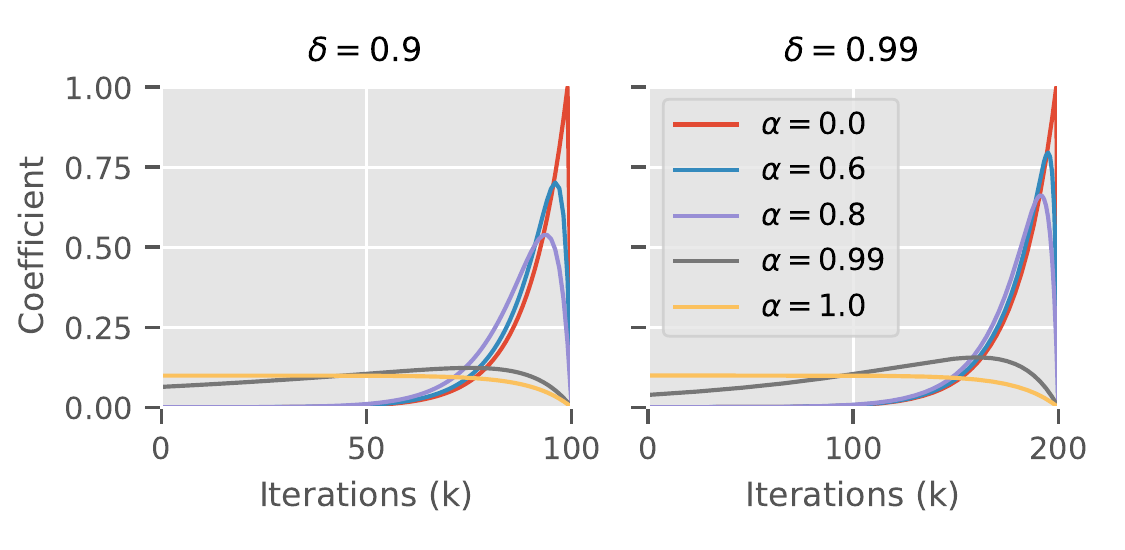}
	\caption{Error decay of GRAPE. Lines show the coefficient \eqref{eq:coefficient determining error decay} with various $\alpha$ as in the legend. $\delta$ is shown on top of each panel. A quantitatively similar result is obtained for different $\delta$. Note that the horizontal axis is $k$, which appears in the exponents of $\alpha$, rather than $K$, which appears in $A_K$.}\label{fig:error_decay}
\end{figure}

Maximum noise-tolerance is obtained when $\alpha=1$. However, there are two issues. First, as argued below, effects of non-noise errors in early iterations linger. In early iterations, an agent tends to explore the limited subset of the state space. As a result, errors are expected to be non-stochastic. Second, the decay rate of $\| V^\pi - \op{\pi} \Psi_0 \|$ is very slow. Indeed, it is $O \left( K^{-1} \right)$ when $\alpha=1$ while it is $O \left( \max \{ \alpha, \delta \}^K \right)$ when $\alpha \neq 1$.

Finally, we argue what happens if $\varepsilon_k (s, a)$ are not noise, and averaging has no effect. Then, using the triangle inequality, we have
\begin{align*}
	&\left\| A^\pi - \Phi_K / A_K \right\|\\
	&\leq o \left( 1 \right) + \frac{2}{A_K} \sum_{k=0}^{K-1} \left\| \varepsilon_k \right\| \sum_{l=0}^{K-k-1} \alpha^{K-k-l-1} \delta^l.
\end{align*}
Thus,
\begin{equation}\label{eq:coefficient determining error decay}
	A_K^{-1} \sum_{l=0}^{K-k-1} \alpha^{K-k-l-1} \delta^l
\end{equation}
determines how quickly effects of past errors decay. Note that $K$ (the number of iterations) is used in $A_K$, and $k$ (an index of iterations) is used in the exponents of $\alpha$. 

Figure~\ref{fig:error_decay} visualize the coefficient clearly illustrating enlarged and lessened effect of the past ($k << K$) and recent errors ($k \approx K$) for a large $\alpha$, respectively. A quantitatively similar result is obtained for other $\delta$.

From Fig.~\ref{fig:error_decay}, one may wonder whether the net effect of errors might be large in GRAPE. To see that this is not the case, let us suppose for simplicity that $\left\| \varepsilon_{K-k} \right\| \leq \varepsilon$, and that $\alpha=1$, which must show a drastic difference from a case with $\alpha=0$. Then,
\begin{align*}
	\lim_{K \rightarrow \infty} \left\| A^\pi - \Phi_K / A_K \right\| \leq \frac{2 \varepsilon}{1 - \delta}.
\end{align*}
The same asymptotic bound is obtained when $\alpha = 0$; thus, the net effect of errors is unchanged.

\section{NUMERICAL EXPERIMENTS}
We conducted numerical experiments to compare GRAPE, Retrace and Retrace with learning rates. We first carried out experiments in finite state-action environments with a tabular representation of functions. The focus of those experiments are confirming the noise-tolerance of GRAPE under a model-free setting. Furthermore, we implemented GRAPE combined with an actor-critic using neural networks and observed its promising performance in benchmark control tasks.

\begin{figure}[t]
	\centering
	\includegraphics[width=\columnwidth]{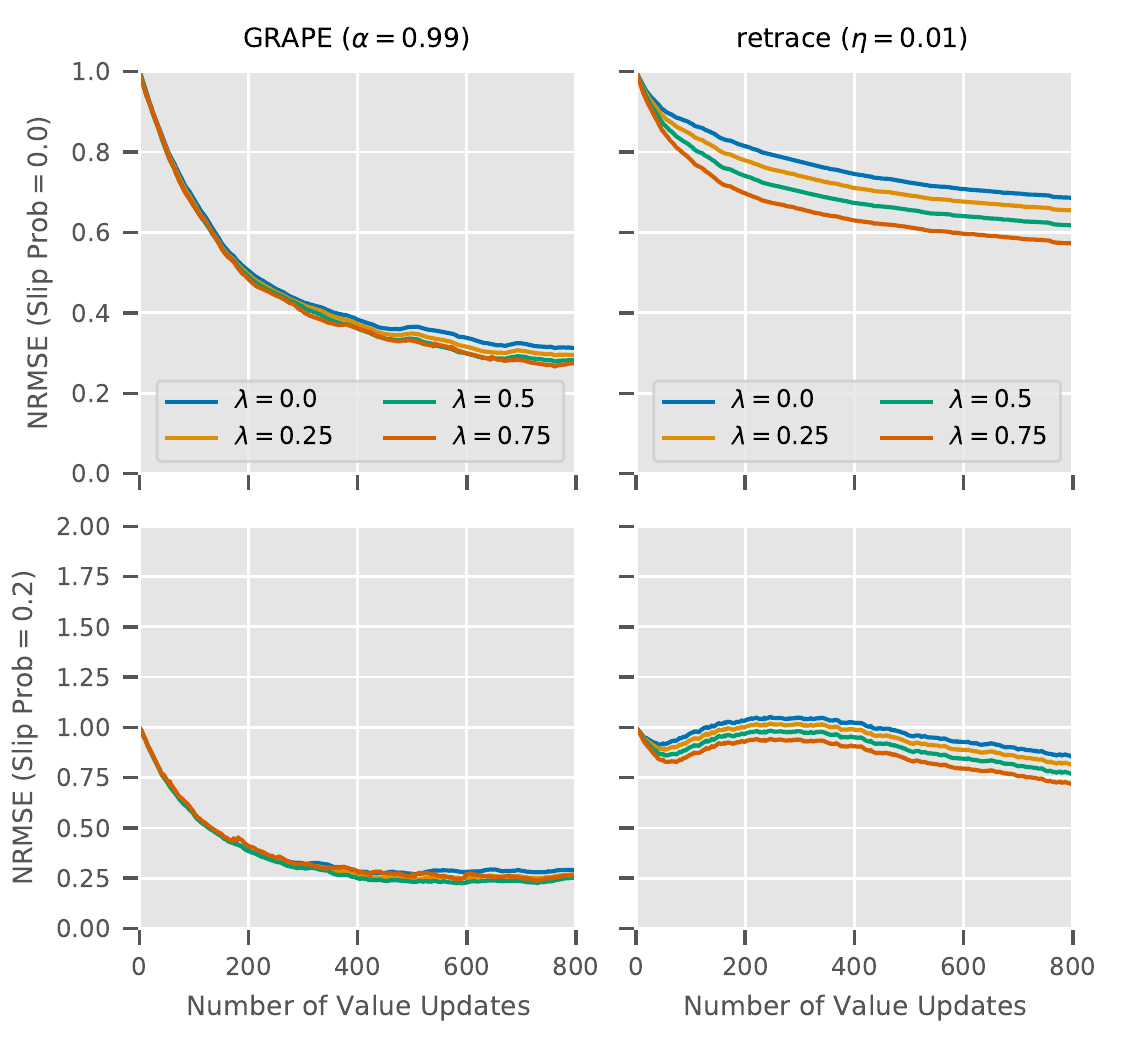}
	\caption{Policy evaluation performance comparison of GRAPE and Retrace with a learning rate in NChain. The figure is same as Fig.~\ref{fig:performance comparison in nchain} except that $\alpha$ of GRAPE and $\eta$ of Retrace are fixed to $0.99$ and $0.01$, respectively. These values are chosen so that the effect of $\lambda$ is most visible.}
	\label{fig:performance comparison in nchain changing lambda}
\end{figure}

\subsection{MODEL-FREE POLICY EVALUATION}
We first carried out model-free policy evaluation experiments in an environment called NChain, which is a larger, stochastic version of an environment in Example~6.2 Random Walk of \citep{sutton_introRL_2018}. The environment is a horizontally aligned linear chain of twenty states in which an agent can move left or right at each time step. However, with a small probability called slip prob ($\leq 0.5$), the agent moves to an opposite direction. The agent can get a small positive reward when it reaches the right end of the chain. The left and right ends of the chain are terminal states.

\begin{figure*}[t!]
	\centering
	\includegraphics[width=\textwidth]{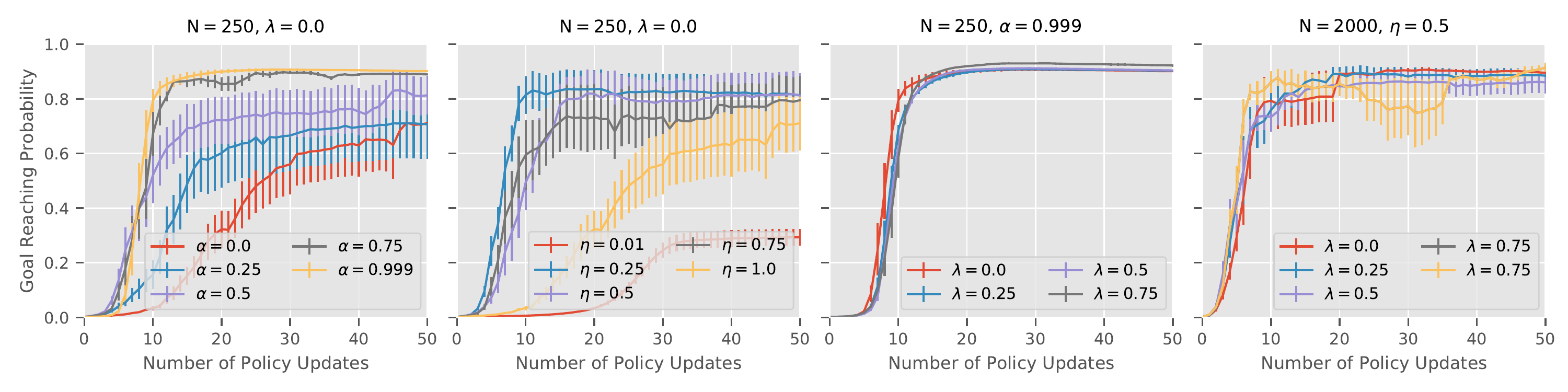}
	\caption{Performance comparison of GRAPE and Retrace with a learning rate in FrozenLake. The horizontal axes show the number of policy updates. The vertical axes show goal-reaching probability (higher is better) computed by DP. An optimal $\alpha$ is almost $1$. The number of samples $N$ used for updating value functions is indicated on top of each panel. Lines show mean performance over six trials. Error bars show standard error. The first and second (from left to right) panels show effects of $\alpha$ in GRAPE and $\eta$ in Retrace with a learning rate, respectively. $N$ and $\lambda$ are fixed to $250$ and $0$, respectively, as shown on top of those panels. The third panel shows the effect of $\lambda$ in GRAPE with $\alpha=0.999$. The last panel shows the effect of $\lambda$ in Retrace with a learning rate $\eta=0.5$ when $N=2,000$. ($\eta=0.5$ has performed best when $N=2,000$ in contrast to a case $N=250$.)}
	\label{fig:performance comparison in frozenlake}
\end{figure*}

The experiments are conducted as follows: one trial consists of $200,000$ interactions, i.e. time steps, of an agent with an environment. At each time step, the agent takes an action $a \sim \mu (\cdot | x)$ given a current state $x$. Then, it observes a subsequent state $y$ with an immediate reward $r$. If the state transition is to a terminal state, an episode ends, and the agent starts again from a random initial state. The interactions are divided into multiple blocks. One block consists of $N = 250$ time steps. After each block, the agent update its value function using $N$ samples of the state transition data $(x, a, r, y, \mu (a|x), d)$ in the block, where $d=1$ if the transition is to a terminal state otherwise $0$. After each block, the agent is reset to the start state. $\Psi_0$ is initialized to $\Psi_0 (x, a)$. $\pi (\cdot | x)$ and $\mu (\cdot | x)$ are sampled from $|\set{A}|$-dimensional Dirichlet distribution with all concentration parameters set to $1$. The discount factor is $0.99$, and $\lambda$ is varied.\footnote{We did the same experiments in FrozenLake. However, no algorithm worked, probably because such a randomly constructed behavior policy $\mu$ hardly reaches a goal, and thus, initial function $\Psi_0 (x, a) = 0$ is already close to the true value.} More implementation details can be found in Appendix~\ref{appendix:frozenlake details}.

Figure~\ref{fig:performance comparison in nchain} visually compares GRAPE and Retrace with a learning rate. $\lambda$ is set to $0$. In all panels, there is a clear tendency that increasing either $\alpha$ or $\eta$ leads to decreased NRMSE, except $\eta = 0.01$. Asymptotic NRMSE of GRAPE and Retrace closely match when $\alpha = 1-\eta$. We note that GRAPE with $\alpha = 0.99$ shows strong noise-tolerance with reasonably fast learning. Because the number of samples in one update is fixed, this result shows significantly more efficient learning by GRAPE. Due to page limitations, we omit experimental results in which the number of samples in one update is $N = 2000$. However, we note that GRAPE with a frequent update with $N=250$ worked better in terms of the number of samples, in accordance with our theory.

Figure~\ref{fig:performance comparison in nchain changing lambda} illustrates the effect of changing $\lambda$. In GRAPE, there is a slight improvement by increasing $\lambda$, whereas in Retrace, there is a clear tendency that increasing $\lambda$ improves learning. A possible reason implied by our theory is that $\delta$ is much smaller than $\alpha = 0.99$; thus, the convergence rate is almost determined by $\alpha$.

\subsection{MODEL-FREE CONTROL}

Next, we carried out model-free control experiments in FrozenLake to investigate the usefulness of GRAPE. The experimental settings are similar to those for the model-free policy evaluation task. Differences are the following: one trial consists of $5,000,000$ interaction time steps. In contrast to a model-free policy evaluation task, there is no block. At each time step, the agent takes an action $a \sim \pi_k (\cdot | x)$, which is repeatedly updated through the trial. The state transition data $(x, a, r, y, \pi_k (a|x), d)$ are stored in a buffer $\set{D}$, of size $500,000$. Every $N = \{250, 2000\}$ (fixed through the trial) time steps, the agent updates its value function using $N$ contiguous samples from the buffer $\set{D}$. Every $100,000$ time steps, the agent updates its policy according to a rule explained below. $\beta \in \{0.1, 0.2, 0.5, 1, 2, 5, 10, 20, 50, 100\}$ are tried for each parameter set $(\alpha, \lambda, N)$ (or $(\eta, \lambda, N)$ when Retrace with a learning rate is used), and we selected one that yielded the highest asymptotic performance. $\Psi_0$ is initialized to $\Psi_0 (x, a)$. $\pi_0$ and $\mu_0$ are initialized to $\pi_0 (a|x) = \mu_0 (a|x) = 1 / |\set{A}|$. More implementation details can be found in Appendix~\ref{appendix:frozenlake details}.

For policy improvement, we used a simple variant of Trust Region Policy Optimization (TRPO) \citep{schulman_trpo_2015}. Its policy updates are given by
\begin{gather}
	\pi_{k+1} (a|x) = \frac{\pi_k (a|x) \exp \left( \beta A^{\pi_k} (x, a) \right)}{\sum_{b \in \set{A}} \pi_k (b|x) \exp \left( \beta A^{\pi_k} (x, b) \right)},\label{eq:TRPO-like policy update}
\end{gather}
with $\pi_0 (a|x) = 1 / |\set{A}|$, where $\beta \in (0, \infty)$. For the derivation of this update rule, see Appendix~\ref{appendix:derivation of reverse TRPO}. In real implementation, $A^{\pi_k}$ is estimated by each algorithm.

Figure~\ref{fig:performance comparison in frozenlake} shows the result. The first and second (from left to right) panels show effects of $\alpha$ and $\eta$, respectively. It is possible to see a clear tendency of performance increase by increased $\alpha$. Particularly, GRAPE with $\alpha=0.999$ outperforms Retrace with any learning rate. The third panel shows the effect of $\lambda$ in GRAPE with $\alpha=0.999$. A slightly better asymptotic performance is seen for $\lambda = 0.75$. However, its effect is not clear. The last panel shows the effect of $\lambda$ in Retrace with a learning rate $\eta=0.5$ when $N=2,000$. ($\eta=0.5$ performed best when $N=2,000$ in contrast to a case $N=250$.) In this case, when $\lambda$ is either $0$ or $0.25$, Retrace's asymptotic performance matches that of GRAPE with $\alpha=0.999$. However, note that eight times more data are used in one update. Moreover, the learning of Retrace with a learning rate is unstable compared to that of GRAPE with $\alpha = 0.999$.

\subsection{GRAPE WITH NEURAL NETWORKS}

GRAPE can be also used for value updates in actor-critic algorithms. Here we show an implementation of actor-critic algorithm combining GRAPE with advantage policy gradient with neural networks. We call it as AC-GRAPE, which can deal with control problems in continuous state space. Details of the implementation can be found in Appendix~\ref{appendix:ac}.

We performed experiments with AC-GRAPE in ``Pendulum-v0'' and ``Acrobot-v1'' environments from OpenAI Gym \citep{openai_gym}, and compared the result with that using Retrace for value updates (AC-Retrace). For Pendulum, we discretized the action space to 15 actions log-uniformly between $-2$ and $2$. In noisy case we added Gaussian white noise to the original observations, where the standard deviation is 0.1 for Pendulum and 0.25 for Acrobot. For both Pendulum and Acrobot, we used discount factor $\gamma=0.99$. Size of the experience buffer was set to 50000. Every 1000 time steps, we conducted a so-called ``test phase'', during which an agent is evaluated for 10 episodes while halting the training. Hyper-parameters for AC-GRAPE and AC-Retrace are determined by grid search.

\begin{figure}[t]
	\centering
	\includegraphics[width=0.5\textwidth]{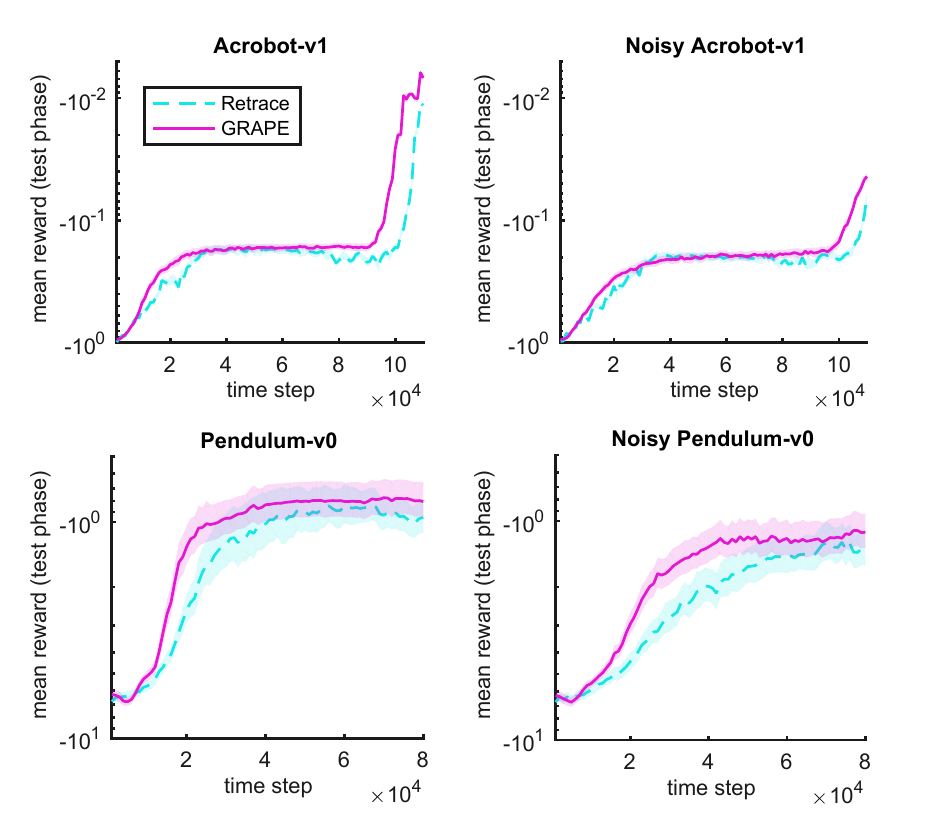}
	\caption{Experimental results of AC-GRAPE in benchmarking control tasks, where the purple curves show performance of the best hyper-parameter setting using GRAPE to obtain the target of $Q$, and the cyan, dashed curves show performance of the best hyper-parameter setting using Retrace to obtain the target of $Q$. Shaded areas indicate 95\% confidence interval.}
	\label{fig:cc}
\end{figure}

Figure~\ref{fig:cc} shows the performance curves, measured by mean rewards in each test phase, of GRAPE and Retrace, with the best hyper-parameter setting. The results showed that in the actor-critic implementations, GRAPE can outperform Retrace.

\section{RELATED RESEARCH}
A line of research most closely related to GRAPE is \citep{azar_dpp_2012,rawlik_thesis_2013,bellemare_action_gap_2016,kozuno_cvi_2017}, in which gap-increasing single-stage lookahead control algorithms are proposed and analyzed. Those papers imply noise-tolerance of gap-increasing operators. However, policy evaluation version of those algorithms is not argued in detail in those papers.

Another line of similar research is off-policy multi-stage lookahead policy evaluation algorithms, such as Retrace \citep{munos_retrace_2016} and tree-backup \citep{tree_backup}. In this paper, we combined the idea of Retrace into our GRAPE algorithm. However, it is straightforward to extend our gap-increasing policy evaluation algorithm to tree-backup-like algorithms.

Our theoretical analysis is similar to those in \citep{munos05,munos07,FarahmandThesis,scherrer2012_nips_on_the_use_of_nonstationary}. However, we did not show $l_p$-norm error bounds due to page limitations although it is not difficult.

\section{CONCLUSION}
In the present paper, we proposed a new policy evaluation algorithm called GRAPE. GRAPE is shown to be efficient and noise-tolerant by both theoretical analysis and experimental evidence. GRAPE has been compared to a state-of-the-art policy evaluation algorithm called Retrace. GRAPE demonstrated significant gains in performance and stability.

Though our theoretical analysis is valid even for continuous action space, we only tested GRAPE in environments with a finite action space. Extending GRAPE to a continuous action space is an important research direction. 

\subsubsection*{Acknowledgments}
This work was supported by JSPS KAKENHI Grant Numbers 16H06563. We thank Dr. Steven D. Aird at Okinawa Institute of Science and Technology for editing and proofreading the paper. We are also grateful to reviewers for valuable comments and suggestions.

\subsubsection*{References}

\bibliographystyle{rusnat}

\begingroup
\renewcommand{\section}[2]{}
\bibliography{uai2019}
\endgroup

\newpage
\appendix

\onecolumn

\section{Notations in Proofs}
For brevity, we use notations different from the main paper. In particular, we use matrix and operator notation. For a function $f$ over a finite set $\set{S}$, $\vec{f}$ denotes an $|\set{S}|$-dimensional vector consisting of $f(s), s \in \set{S}$. $\vec{f}(s)$ denotes $f(s)$. We let $\set{Q}$ and $\set{V}$ denote sets of $\left| \set{X} \times \set{A} \right|$ and $\left| \set{X} \right|$ dimensional vectors, respectively. For a policy $\pi$, $\op{\pi}: \set{Q} \rightarrow \set{V}$ denotes a matrix such that $\left( \op{\pi} \vec{Q} \right) (x, a) := \sum_{a \in \set{A}} \pi (a|x) Q (x, a)$. A matrix $\op{P}: \set{V} \rightarrow \set{Q}$ is defined such that $\left( \op{P} \vec{V} \right) (x, a) := \sum_{y \in \set{X}} P (y|x, a) V (x)$, where $P (y| x, a) := \int_{\R} P(y, r|x, a)\, dr$ . A matrix $\op{P^\pi}$ is defined by $\op{P} \op{\pi} $. The Bellman operator $\op{T^\pi}$ is an operator such that $\op{T^\pi} \vec{Q} := \vec{r} + \gamma \op{P^\pi} \vec{Q}$, where $r (x, a)$ is an expected reward $\sum_{y \in \set{X}, r \in \set{R}} r P_0 (y, r | x, a)$. Note that by extending $V \in \set{V}$ by $V (x, a) := V(x)$, we can regard $V$ as an element of $\set{Q}$. Thus, the addition and subtraction of $V \in \set{V}$ and $Q \in \set{Q}$ are naturally defined as, for example, $( \vec{Q} - \vec{V} )(x, a) := Q(x, a) - V(x)$. Similarly, when an $|\set{X}||\set{A}| \times |\set{X}||\set{A}|$ matrix, say $\op{M_1}$, is added to an $|\set{X}| \times |\set{X}||\set{A}|$ matrix, say $\op{M_2}$, we extend $\op{M_2}$ to an $|\set{X}||\set{A}| \times |\set{X}||\set{A}|$ matrix such that $\left( \op{M_2} \vec{Q} \right) (x, a) := \left( \op{M_2} \vec{Q} \right) (x)$. For an operator $\op{O}$, we define its $k$-th power $\op{O}^k$, where $k \in \{0, 1, \ldots\}$, such that $\op{O}^k \vec{f} := \op{O}^{k-1} \left( \op{O} \vec{f} \right) := \cdots$ with $\op{O}^0 \vec{f} := \op{I} \vec{f} = \vec{f}$. For any policy $\pi$, a matrix $(\op{I} - \kappa \op{P^\pi})^{-1}$ is well defined as long as $\kappa \in [0, 1)$ and given as $\sum_{t=0}^\infty \kappa^t (\op{P^\pi})^t$. Similarly, a matrix $(\op{I} - \kappa \op{P^{c \mu}})^{-1}$ is well defined and given as $\sum_{t=0}^\infty \kappa^t (\op{P^{c \mu}})^t$.

\section{Proof of Theorem~\ref{theorem:convergence of Psi_k} and \ref{theorem:agrape performance bound}}\label{appendix:proof}
Because we use some lemmas here later in proofs of Theorem~\ref{theorem:agrape performance bound}, we consider approximate GRAPE updates \eqref{eq:agrape update}.

We first prove Lemma~\ref{lemma:G is a contraction} that shows the contraction property of $\op{G^{c \mu}_\lambda}$.
\begin{proof}[Proof of Lemma~\ref{lemma:G is a contraction}]
	Indeed,
	\begin{align*}
	\vec{Q^\pi} - \op{G^{c \mu}_\lambda} \vec{Q}
	&= \gamma \op{P^{\pi}} \vec{\Delta} + \gamma \lambda \left( \op{I} - \gamma \lambda \op{P^{c \mu}} \right)^{-1} \op{P^{\pi}} \left( \gamma \op{P^{\pi}} - \op{I} \right) \vec{\Delta}\\
	&= \gamma \op{P^{\pi}} \vec{\Delta} + \gamma \lambda \left( \op{I} - \gamma \lambda \op{P^{c \mu}} \right)^{-1} \left( \gamma \op{P^{\pi}} - \gamma \lambda \op{P^{c \mu}} + \gamma \lambda \op{P^{c \mu}} - \op{I} \right) \op{P^{\pi}} \vec{\Delta}\\
	&= \gamma (1 - \lambda) \op{P^{\pi}} \vec{\Delta} + \gamma^2 \lambda \left( \op{I} - \gamma \lambda \op{P^{c \mu}} \right)^{-1} \left(\op{P^{\pi}} - \lambda \op{P^{c \mu}} \right) \op{P^{\pi}} \vec{\Delta},
	\end{align*}
	where we used a shorthand notation $\vec{\Delta} := \vec{Q^\pi} - \vec{Q}$. Therefore, applying the triangular inequality and simply noting an operator norm of $\left\| \op{P^{\pi}} \right\| := \max_{\vec{Q}, \|\vec{Q}\| = 1} \left\| \op{P^{\pi}} \vec{Q} \right\| = 1$,
	\begin{align*}
	&\left\| \vec{Q^\pi} - \op{G^{c \mu}_\lambda} \vec{Q} \right\|
	\leq \gamma (1 - \lambda) \left\| \vec{\Delta} \right\| + \gamma^2 \lambda \left\| \left( \op{I} - \gamma \lambda \op{P^{c \mu}} \right)^{-1} \left(\op{P^{\pi}} - \lambda \op{P^{c \mu}} \right) \right\| \left\| \vec{\Delta} \right\|.
	\end{align*}
	Munos et al. showed that $\| \left( \op{I} - \gamma \lambda \op{P^{c \mu}} \right)^{-1} \left(\op{P^{\pi}} - \lambda \op{P^{c \mu}} \right) \| \leq 1$ \citep{munos_retrace_2016}. Thus, $\left\| \vec{Q^\pi} - \op{G^{c \mu}_\lambda} \vec{Q} \right\| \leq \gamma (1 - \lambda) \left\| \vec{\Delta} \right\| + \gamma^2 \lambda \left\| \vec{\Delta} \right\| = \delta \left\| \vec{\Delta} \right\|$.
\end{proof} 

\begin{remark}
	As is seen in the proof, $\left\| \vec{Q^\pi} - \op{G^{c \mu}_\lambda} \vec{Q} \right\|
	\leq \delta \left\| \op{\pi} \vec{\Delta} \right\|$ holds too.
\end{remark}

We next prove the following lemma that relates $\Psi_k$ with $\Psi_0$.
\begin{lemma}\label{lemma:simplified Psi_k anoter expression}
	Let $\op{H^{c \mu}_\lambda}: \set{Q} \rightarrow \set{Q}$ denote an operator that maps $\vec{Q}$ to $\op{H^{c \mu}_\lambda} \vec{Q} := \gamma \op{P^\pi} \vec{Q} + \left( \op{I} - \gamma \lambda \op{P^{c\mu}} \right)^{-1} \op{P^{\pi}} \left( \gamma \op{P^{\pi}} - \op{I} \right) \vec{Q}$. For any positive integer $K$, $\Psi_{K}$ of GRAPE can be rewritten as
	\begin{align*}
		\vec{\Psi_{K}} = A_{K} \vec{q_{K}} - \alpha A_{K-1} \op{\pi} \vec{q_{K-1}},
	\end{align*}
	where $\vec{q_0} = \vec{\Psi_0}$,
	\begin{equation*}
		A_K \vec{q_K} := \sum_{k=1}^{K} \alpha^{K-k} \left( \op{G^{c \mu}_\lambda} \right)^{k} \vec{\Psi_0} + \sum_{k=0}^{K-1} \left( \op{H^{c \mu}_\lambda} \right)^{K-k-1} \vec{E_k},
	\end{equation*}
	$A_0 := 0$, $A_K = \sum_{k=0}^{K-1} \alpha^k$, and $\vec{E_k} := \sum_{l=0}^{k} \alpha^l \vec{\varepsilon_{k-l}}$.
\end{lemma}

\begin{proof}[Proof of Lemma~\ref{lemma:simplified Psi_k anoter expression}]
	Note that
	\begin{align*}
		\op{G^{c \mu}_\lambda} \left( \vec{Q} + \vec{R} \right)
		= \op{T^{\pi}} \vec{Q} + \gamma \op{P^{\pi}} \vec{R} + \gamma \lambda \left( \op{I} - \gamma \lambda \op{P^{c \mu}} \right)^{-1} \op{P^{\pi}} \left( \op{T^{\pi}} \vec{Q} + \gamma \op{P^{\pi}} \vec{R} - \vec{Q} - \vec{R} \right)
		= \op{G^{c \mu}_\lambda} \vec{Q} + \op{H^{c \mu}_\lambda} \vec{R}.
	\end{align*}
	Because $\op{\pi} \vec{\Phi_k} = 0$ for any $k$, it follows that $\op{P^\pi} \vec{\Phi_k} = 0$. From this, we have
	\begin{align}\label{eq:G Psi_k to G Psi_0}
		\op{G^{c \mu}_\lambda} \vec{\Psi_{K-1}}
		&= \op{G^{c \mu}_\lambda} \left( \op{G^{c \mu}_\lambda} \vec{\Psi_{K-2}} + \alpha \vec{\Phi_{K-2}} + \vec{\varepsilon_{K-2}} \right)\nonumber\\
		&= \op{G^{c \mu}_\lambda} \op{G^{c \mu}_\lambda} \vec{\Psi_{K-2}} + \op{H^{c \mu}_\lambda} \vec{\varepsilon_{K-2}}\nonumber\\
		&\vdots\nonumber\\
		&= \left( \op{G^{c \mu}_\lambda} \right)^K \vec{\Psi_0} + \sum_{k=1}^{K-1} \left( \op{H^{c \mu}_\lambda} \right)^{K-k-1} \vec{\varepsilon_k}.
	\end{align}
	Similarly, $\op{\pi} \vec{\Psi_{K-1}} = \op{\pi} \op{G^{c \mu}_\lambda} \vec{\Psi_{K-2}} + \op{\pi} \vec{\varepsilon_{K-2}} = \op{\pi} \left( \op{G^{c \mu}_\lambda} \right)^{K-1} \vec{\Psi_0} + \op{\pi} \sum_{k=0}^{K-2} \left( \op{H^{c \mu}_\lambda} \right)^{K-k-2} \vec{\varepsilon_k}$. As a result,
	\begin{align}\label{eq:simplified Phi_k another expression}
		\vec{\Phi_{K-1}}
		&= \op{G^{c \mu}_\lambda} \vec{\Psi_{K-2}} + \alpha \vec{\Phi_{K-2}} + \vec{\varepsilon_{K-2}} - \op{\pi} \vec{\Psi_{K-1}}\nonumber\\
		&= \alpha \vec{\Phi_{K-2}} + \left( \op{I} - \op{\pi} \right) \left[ \left( \op{G^{c \mu}_\lambda} \right)^{K-1} \vec{\Psi_0} + \sum_{k=0}^{K-2} \left( \op{H^{c \mu}_\lambda} \right)^{K-k-2} \vec{\varepsilon_k} \right]\nonumber\\
		&= \alpha^2 \vec{\Phi_{K-3}} + \left( \op{I} - \op{\pi} \right) \sum_{l=0}^1 \alpha^l \left[ \left( \op{G^{c \mu}_\lambda} \right)^{K-l-1} \vec{\Psi_0} + \sum_{k=0}^{K-l-2} \left( \op{H^{c \mu}_\lambda} \right)^{K-k-l-2} \vec{\varepsilon_k} \right]\nonumber\\
		&\vdots\nonumber\\
		&= \left( \op{I} - \op{\pi} \right) \sum_{l=0}^{K-2} \alpha^l \left[ \left( \op{G^{c \mu}_\lambda} \right)^{K-l-1} \vec{\Psi_0} + \sum_{k=0}^{K-l-2} \left( \op{H^{c \mu}_\lambda} \right)^{K-k-l-2} \vec{\varepsilon_k} \right]\nonumber\\
		&= \left( \op{I} - \op{\pi} \right) \sum_{k=1}^{K-1} \alpha^{K-k-1} \left( \op{G^{c \mu}_\lambda} \right)^{k} \vec{\Psi_0} + \left( \op{I} - \op{\pi} \right) \sum_{k=0}^{K-2} \left( \op{H^{c \mu}_\lambda} \right)^{K-k-2} \vec{E_k}.
	\end{align}
	Because $\vec{\Psi_K} = \op{G^{c \mu}_\lambda} \vec{\Psi_{K-1}} + \alpha \vec{\Phi_{K-1}} + \vec{\varepsilon_{K-1}}$, it follows that
	\begin{align*}
		\vec{\Psi_{K}}
		&= \left( \op{G^{c \mu}_\lambda} \right)^{K} \vec{\Psi_0} + \sum_{k=0}^{K-1} \left( \op{H^{c \mu}_\lambda} \right)^k \vec{\varepsilon_{K-k-1}} + \left( \op{I} - \op{\pi} \right) \sum_{k=1}^{K-1} \alpha^{K-k} \left( \op{G^{c \mu}_\lambda} \right)^{k} \vec{\Psi_0} + \alpha \left( \op{I} - \op{\pi} \right) \sum_{k=0}^{K-2} \left( \op{H^{c \mu}_\lambda} \right)^{K-k-2} \vec{E_k}\\
		&= \sum_{k=1}^{K} \alpha^{K-k} \left( \op{G^{c \mu}_\lambda} \right)^{k} \vec{\Psi_0} + \sum_{k=0}^{K-1} \left( \op{H^{c \mu}_\lambda} \right)^{K-k-1} \vec{E_k} - \alpha \op{\pi} \left[ \sum_{k=1}^{K-1} \alpha^{K-k-1} \left( \op{G^{c \mu}_\lambda} \right)^{k} \vec{\Psi_0} + \sum_{k=0}^{K-2} \left( \op{H^{c \mu}_\lambda} \right)^{K-k-2} \vec{E_k} \right]\\
		&= A_K \vec{q_K} - \alpha A_{K-1} \op{\pi} \vec{q_{K-1}}.
	\end{align*}
	This concludes the proof.
\end{proof}

By combining Lemma~\ref{lemma:G is a contraction} and \ref{lemma:simplified Psi_k anoter expression}, the following lemma is readily proven.
\begin{lemma}\label{lemma:convergence of q_k}
	If $\varepsilon_k (x, a) = 0$ for any $k \in \{0, 1, \ldots, K\}$, $(x, a) \in \set{X} \times \set{A}$, then, $q_K$ uniformly converges to $Q^{\pi}$ with convergence rates $O(\max\{\alpha, \delta\}^K)$ when $\alpha \neq 1$ and $O(K^{-1})$ when $\alpha = 1$.
\end{lemma}

\begin{proof}
	Note that $\vec{Q^\pi} = A_K^{-1} \sum_{k=1}^{K} \alpha^{K-k} \vec{Q^\pi}$, and that $A_K \vec{q_K} = \sum_{k=1}^{K} \alpha^{K-k} \left( \op{G^{c \mu}_\lambda} \right)^{k} \vec{\Psi_0}$. Because $\op{G^{c \mu}_\lambda}$ is a contraction,
	\begin{align*}
		\left\| \vec{Q^{\pi}} - \vec{q_K} \right\|
		\leq \dfrac{1}{A_K} \sum_{k=1}^{K} \alpha^{K-k} \left\| \vec{Q^{\pi}} - \left( \op{G^{c \mu}_\lambda} \right)^k \vec{\Psi_0} \right\|
		\leq \dfrac{1}{A_K} \sum_{k=1}^{K} \alpha^{K-k} \delta^k \left\| \vec{V^{\pi}} - \op{\pi} \vec{\Psi_0} \right\|.
	\end{align*}
	It is lengthy to explain how the last inequality is derived. However, the derivation is intuitively understood by looking at a case where $\lambda = 0$. In that case,
	\begin{align*}
	    \left\| \vec{Q^{\pi}} - \op{G^{c \mu}_\lambda} \vec{\Psi_0} \right\|
	    = \gamma \left\| \op{P} \op{\pi} \vec{Q^{\pi}} - \op{P} \op{\pi} \vec{\Psi_0} \right\| \leq \gamma \left\| \op{\pi} \vec{Q^{\pi}} - \op{\pi} \vec{\Psi_0} \right\|.
	\end{align*}
    It is straightforward to extend this discussion for a general $\lambda$. We can rewrite the coefficient of $\left\| \vec{V^{\pi}} - \op{\pi} \vec{\Psi_0} \right\|$ as
	\begin{equation*}
		\dfrac{1}{A_K} \sum_{k=1}^{K} \alpha^{K-k} \delta^k = \frac{\delta}{A_K} \frac{\alpha^K - \delta^K}{\alpha - \delta}.
	\end{equation*}
	Accordingly, the convergence rate is given by $O(\max\{\alpha, \delta\}^K)$ when $\alpha \neq 1$ and $O(K^{-1})$ when $\alpha = 1$.
\end{proof}

Theorem~\ref{theorem:convergence of Psi_k} is an immediate consequence of Lemma~\ref{lemma:convergence of q_k}. Indeed, for example, note that
\begin{gather*}
	\left\| A^{\pi} - \frac{1}{A_K} \vec{\Phi_K} \right\| = \left\| \vec{q_K} - \op{\pi} \vec{q_K} - \vec{A^{\pi}} \right\| \leq \dfrac{2}{A_K} \sum_{k=1}^{K} \alpha^{K-k} \delta^k \left\| \vec{V^{\pi}} - \op{\pi} \vec{\Psi_0} \right\|.
\end{gather*}
Thus, we have proven Theorem~\ref{theorem:convergence of Psi_k}.

Theorem~\ref{theorem:agrape performance bound} is proven by noting that $\vec{\Phi_K} / A_K = \vec{q_K} - \op{\pi} \vec{q_K}$ from Eq.~\eqref{eq:simplified Phi_k another expression}, and thus,
\begin{align*}
	\left\| \vec{A^{\pi}} - \frac{1}{A_K} \vec{\Phi_K} \right\|
	\leq \left\| \vec{q_K} - \vec{Q^{\pi}} \right\|
	\leq \dfrac{2}{A_K} \sum_{k=1}^{K} \alpha^{K-k} \delta^k \left\| \vec{V^{\pi}} - \op{\pi} \vec{\Psi_0} \right\| + \frac{2}{A_K} \sum_{k=0}^{K-1} \delta^{K-k-1} \left\| \vec{E_k} \right\|,
\end{align*}
where we used the following lemma that shows $\op{H^{c \mu}_\lambda}$ is a contraction with modulus $\delta$:
\begin{lemma}\label{lemma:H is a contraction}
	$\op{H^{c \mu}_\lambda}$ is a contraction around $\vec{0}$ with modulus $\delta$.
\end{lemma}

\begin{proof}
	Indeed
	\begin{align*}
		\op{H^{c \mu}_\lambda} \vec{Q}
		&= \gamma \op{P^{\pi}} \vec{Q} + \gamma \lambda \left( \op{I} - \gamma \lambda \op{P^{c \mu}} \right)^{-1} \op{P^{\pi}} \left( \gamma \op{P^{\pi}} - \op{I} \right) \vec{Q},
	\end{align*}
	and thus, by a discussion similar to the proof of Lemma~\ref{lemma:G is a contraction}, we conclude that $\op{H^{c \mu}_\lambda}$ is a contraction around $\vec{0}$ with modulus $\delta$.
\end{proof}

\section{Discussion on Remark~\ref{remark:on reuse of Psi}}\label{appendix:reuse of Psi}
We prove the inequality~\eqref{eq:reuse of Psi}. Indeed,
\begin{align*}
    \left\| \vec{V^\pi} - \op{\pi} \vec{\Psi_0} \right\|
    &= \left\| \vec{V^\pi} - \op{\pi} \vec{Q^{\widetilde{\pi}}} + \op{\pi} \vec{Q^{\widetilde{\pi}}} - \vec{V^{\widetilde{\pi}}} + \vec{V^{\widetilde{\pi}}} - \op{\widetilde{\pi}} \vec{\Psi_0} + \op{\widetilde{\pi}} \vec{\Psi_0} - \op{\pi} \vec{\Psi_0} \right\|\\
    &\leq \left\| \vec{V^\pi} - \op{\pi} \vec{Q^{\widetilde{\pi}}} \right\| + \left\| \op{\pi} \vec{Q^{\widetilde{\pi}}} - \vec{V^{\widetilde{\pi}}} \right\| + \left\| \vec{V^{\widetilde{\pi}}} - \op{\widetilde{\pi}} \vec{\Psi_0} \right\| + \left\| \op{\widetilde{\pi}} \vec{\Psi_0} - \op{\pi} \vec{\Psi_0} \right\|.
\end{align*}
Thus, we need upper bounds for the first, second and fourth terms. We drive them one by one.

Consider $\left\| \vec{V^\pi} - \op{\pi} \vec{Q^{\widetilde{\pi}}} \right\|$. We have $\left\| \vec{V^\pi} - \op{\pi} \vec{Q^{\widetilde{\pi}}} \right\| \leq \left\| \vec{Q^\pi} - \vec{Q^{\widetilde{\pi}}} \right\|$. By a bit of linear algebra,
\begin{align*}
    \vec{Q^\pi} - \vec{Q^{\widetilde{\pi}}}
    &= \gamma \op{P^{\pi}} \left( \vec{Q^\pi} - \vec{Q^{\widetilde{\pi}}} \right) + \gamma \left( \op{P^{\widetilde{\pi}}} - \op{P^\pi} \right) \vec{Q^{\widetilde{\pi}}}\\
    &= \gamma \left( \op{I} - \gamma \op{P^{\pi}} \right)^{-1} \left( \op{P^{\widetilde{\pi}}} - \op{P^\pi} \right) \vec{Q^{\widetilde{\pi}}}
    = = \gamma \left( \op{I} - \gamma \op{P^{\pi}} \right)^{-1} \op{P} \left( \op{\widetilde{\pi}} - \op{\pi} \right) \vec{Q^{\widetilde{\pi}}}.
\end{align*}
Thus,
\begin{align*}
    \left\| \vec{Q^\pi} - \vec{Q^{\widetilde{\pi}}} \right\| \leq \frac{\gamma}{1 - \gamma} \max_{x \in \set{X}, a \in \set{A}} \left| (\pi (a|x) - \widetilde{\pi} (a|x)) Q^{\widetilde{\pi}} \right| \leq \frac{\gamma}{1 - \gamma} V_{max} \max_{x \in \set{X}, a \in \set{A}} \left| \pi (a|x) - \widetilde{\pi} (a|x) \right|.
\end{align*}
As $\max_{a \in \set{A}} \left| \pi (a|x) - \widetilde{\pi} (a|x) \right|$ is a total variation, Pinsker's inequality implies $\max_{x \in \set{X}, a \in \set{A}} \left| \pi (a|x) - \widetilde{\pi} (a|x) \right| \leq \sqrt{2} D^{1/2}$.

Next, consider $\left\| \op{\pi} \vec{Q^{\widetilde{\pi}}} - \vec{V^{\widetilde{\pi}}} \right\|$. We have
\begin{align*}
    \left\| \op{\pi} \vec{Q^{\widetilde{\pi}}} - \vec{V^{\widetilde{\pi}}} \right\| = \max_{x \in \set{X}, a \in \set{A}} \left| \left( \pi (a|x) - \widetilde{\pi} (a|x) \right) Q^{\widetilde{\pi}} (x, a) \right| \leq \max_{x \in \set{X}, a \in \set{A}} \left| \pi (a|x) - \widetilde{\pi} (a|x) \right| V_{max} \leq \sqrt{2} D^{1/2} V_{max},
\end{align*}
where we used Pinsker's inequality again.

Finally, consider $\left\| \op{\pi} \vec{Q^{\widetilde{\pi}}} - \vec{V^{\widetilde{\pi}}} \right\|$. We have
\begin{align*}
    \left\| \op{\widetilde{\pi}} \vec{\Psi_0} - \op{\pi} \vec{\Psi_0} \right\| = \max_{x \in \set{X}, a \in \set{A}} \left| \left( \pi (a|x) - \widetilde{\pi} (a|x) \right) \Psi_0 (x, a) \right| \leq \max_{x \in \set{X}, a \in \set{A}} \left| \pi (a|x) - \widetilde{\pi} (a|x) \right| \left\| \vec{\Psi_0} \right\| \leq \sqrt{2} D^{1/2} \left\| \vec{\Psi_0} \right\|,
\end{align*}
where we used Pinsker's inequality again.

In summary, we have
\begin{align*}
    \left\| \vec{V^\pi} - \op{\pi} \vec{\Psi_0} \right\|
    &\leq \frac{ \sqrt{2} \gamma}{1 - \gamma} V_{max} D^{1/2} + \sqrt{2} D^{1/2} V_{max} + \left\| \vec{V^{\widetilde{\pi}}} - \op{\widetilde{\pi}} \vec{\Psi_0} \right\| + \sqrt{2} D^{1/2} \left\| \vec{\Psi_0} \right\|\\
    &= \frac{ \sqrt{2} V_{max}}{1 - \gamma} D^{1/2} + \sqrt{2} D^{1/2} \left\| \vec{\Psi_0} \right\| + \left\| \vec{V^{\widetilde{\pi}}} - \op{\widetilde{\pi}} \vec{\Psi_0} \right\|.
\end{align*}

\section{Discussion on Which Estimator of $\op{G^{c \mu}_\lambda} \Psi_k$ to be Used}\label{sec:discussion on which estimator to be used}
We discuss possible estimators of $\op{G^{c \mu}_\lambda} \vec{\Psi_k}$. For ease of reading, we recall an explicit form of $\op{G^{c \mu}_\lambda} \vec{\Psi_k}$, which is
\begin{align*}
	\op{T^{\pi}} \vec{\Psi} + \sum_{t=0}^\infty \gamma^{t+1} \lambda^{t+1} \left( \op{P^{c \mu}} \right)^t \op{P^{\pi}} \left( \op{T^{\pi}} \vec{\Psi} - \vec{\Psi} \right)
	=r + \gamma \op{P} \op{\pi} \vec{\Psi} + \sum_{t=0}^\infty \gamma^{t+1} \lambda^{t+1} \left( \op{P^{c \mu}} \right)^t \op{P} \left( \op{\pi} \vec{r} + \gamma \op{P} \op{{\pi}} \vec{\Psi} - \op{\pi} \vec{\Psi} \right),
\end{align*}
where we omit the subscript $k$ of $\Psi_k$ to avoid cluttered notation.

First of all, we argue that it is not a good idea to estimate $\sum_{a \in \set{A}} \pi (a|x) \Psi (x, a)$ by $\Psi (x, a), a \sim \pi (\cdot|x)$ or $\rho (x, a) \Psi (x, a), a \sim \mu (\cdot|x)$, where $\rho$ is a importance sampling ratio $\rho (x, a) := \pi (a|x) / \mu (a|x)$. The reason is that the variances of such estimators tend to be high. To confirm it, note that from Lemma~\ref{lemma:simplified Psi_k anoter expression}, we have $\vec{\Psi_k} = A_k \vec{q_k} - \alpha A_{k-1} \op{\pi} \vec{q_{k-1}}$. Furthermore, from Lemma~\ref{lemma:convergence of q_k}, we have $\vec{q_k} \approx \vec{Q^\pi}$. Accordingly, $\vec{\Psi_k} \approx A_k \vec{Q^\pi} - \alpha A_{k-1} \vec{V^\pi}$. Suppose that it holds with equality. Then, the variance of $\Psi_k (x, a), a \sim \pi (\cdot|x)$, for example, is given by
\begin{align*}
	\mathbb{V} \Psi_k (x, \cdot)
	&= \sum_{a \in \set{A}} \pi (a|x) \left[ \left( \Psi_k (x, a) - \sum_{b \in \set{A}} \pi(b | x) \Psi_k (x, b) \right)^2 \right]\\
	&= \sum_{a \in \set{A}} \pi (a|x) \left[ \left( A_k Q^\pi (x, a) - \alpha A_{k-1} V^\pi (x) - V^\pi (x) \right)^2 \right]\\
	&= A_k^2 \sum_{a \in \set{A}} \pi (a|x) A^\pi (x, a)^2.
\end{align*}
Thus, it is proportional to $A_k$, which is large when $\alpha \approx 1$.

Accordingly, one of the most straightforward and reasonable estimator of $\left( \op{G^{c \mu}_\lambda} \Psi_k \right) (x, a)$ is
\begin{align*}
	&r_0 + \gamma \left( \op{\pi} \Psi \right) (x_1) + \gamma \lambda \rho (x_1, a_1) \left[ r_1 + \gamma \left( \op{{\pi}} \Psi \right) (x_2) - \left( \op{\pi} \Psi \right) (x_1) \right]\\
	&\hspace{5em} + \gamma^2 \lambda^2 c (x_1, a_1) \rho (x_2, a_2) \left[ r_2 + \gamma \left( \op{{\pi}} \Psi \right) (x_3) - \left( \op{\pi} \Psi \right) (x_2) \right] + \cdots\\
	&=r_0 + \gamma \left( \op{\pi} \Psi \right) (x_1) + \sum_{t=0}^\infty \gamma^{t+1} \lambda^{t+1} \prod_{u=1}^{t} c (x_u, a_u) \rho (x_{t+1}, a_{t+1}) \left[ r_{t+1} + \gamma \left( \op{{\pi}} \Psi \right) (x_{t+2}) - \left( \op{\pi} \Psi \right) (x_{t+1}) \right],
\end{align*}
where $x_0=x, a_0=a, a_{1:\infty} \sim \mu (\cdot | x_t)$, and $\prod_{u=1}^{0} c (x_u, a_u) = 1$.

We further try to improve the above estimator by using control variates. Let us consider the variance of
\begin{equation*}
	\rho (x_t, a_t) \left[ r_t + \gamma \left( \op{{\pi}} \Psi \right) (x_{t+1}) - \left( \op{\pi} \Psi \right) (x_t) \right],
\end{equation*}
where $x_t$ is given, and $a_t \sim \mu (\cdot | x_t)$. We can add a control variate $\kappa (x_t, a_t)$ to obtain
\begin{equation*}
	\rho (x_t, a_t) \left[ r_t + \gamma \left( \op{{\pi}} \Psi \right) (x_{t+1}) - \left( \op{\pi} \Psi \right) (x_t) - \kappa (x_t, a_t) \right]
\end{equation*}
such that $\sum_{a_t \in \set{A}} \pi (a_t | x_t) \kappa (x_t, a_t) = 0$. Again, suppose that $\vec{\Psi_k} = A_k \vec{Q^\pi} - \alpha A_{k-1} \vec{V^\pi}$. Then, its variance is
\begin{equation*}
	\sum_{a_t \in \set{A}} \mu (a_t | x_t) \rho (x_t, a_t)^2 \left[ \left( A^\pi (x_t, a_t) - \kappa (x_t, a_t) \right)^2 \right].
\end{equation*}
Clearly, $\kappa (x_t, a_t) = A^\pi (x_t, a_t)$ is the best control variate. As an estimate of $A^\pi (x_t, a_t)$ is given by $\Phi_k (x_t, a_t) / A_k \approx (1 - \alpha) \Phi_k (x_t, a_t)$, we replace $\kappa (x_t, a_t)$ with it to obtain an estimator
\begin{equation*}
	\rho (x_t, a_t) \left[ r_t + \gamma \left( \op{{\pi}} \Psi \right) (x_{t+1}) - (1 - \alpha) \Psi (x_t, a_t) - \alpha \left( \op{\pi} \Psi \right) (x_t) \right].
\end{equation*}
In fact, this estimator worked well in experiments.

In summary, we propose to use the following estimator:
\begin{align*}
	r_0 + \gamma \left( \op{\pi} \Psi \right) (x_1) + \sum_{t=0}^\infty \gamma^{t+1} \lambda^{t+1} \prod_{u=1}^{t} c (x_u, a_u) \rho (x_{t+1}, a_{t+1}) \Delta_{t+1},
\end{align*}
where $x_0=x, a_0=a, a_{1:\infty} \sim \mu (\cdot | x_t)$, $\prod_{u=1}^{0} c (x_u, a_u) = 1$, and 
\begin{align*}
	\Delta_t := r_t + \gamma \left( \op{{\pi}} \Psi \right) (x_{t+1}) - (1 - \alpha) \Psi (x_t, a_t) - \alpha \left( \op{\pi} \Psi \right) (x_t).
\end{align*}

\section{Derivation of Policy Update \eqref{eq:TRPO-like policy update}}\label{appendix:derivation of reverse TRPO}
TRPO uses the following policy update rule \citep{schulman_trpo_2015}:
\begin{equation*}
	\pi_{k+1} = \argmax_{\pi} \E_{\substack{x \sim \rho_k\\a \sim \pi}} \left[ A^{\pi_k} (x, a) \right]
	\text{ subject to }
	\E_{\substack{x \sim \rho_k}} \left[ D_{KL} \left( \pi_k (\cdot | x) \middle\| \pi (\cdot | x) \right) \right] \leq \delta,
\end{equation*}
where $\rho_k: \set{X} \rightarrow [0, 1]$ is a state visitation frequency under the policy $\pi_k$, and $D_{KL} \left( \pi_k (\cdot | x) \middle\| \pi (\cdot | x) \right)$ is the KL divergence between $\pi_k (\cdot | x)$ and $\pi (\cdot | x)$. The original theory on which TRPO based states that monotonic policy improvement is guaranteed if the maximum total variation between $\pi_k (\cdot | x)$ and $\pi (\cdot | x)$ is small enough. The KL constraint above is used since KL divergence is an upper bound of total variation. Our policy update use the following update rule
\begin{equation}\label{eq:reverse TRPO problem}
	\pi_{k+1} = \argmax_{\pi} \E_{\substack{x \sim \rho_k\\a \sim \pi}} \left[ A^{\pi_k} (x, a) - \tau D_{KL} \left( \pi (\cdot | x) \middle\| \pi_k (\cdot | x) \right) \right],
\end{equation}
where note that the order of $\pi$ and $\pi_k$ in the KL divergence is reversed, and KL regularizer is used. As we show now, this problem can be analytically solved when $\set{S}$ and $\set{A}$ are finitely countable, and thus, $\pi$ is a $| \set{S} \times \set{A}|$-dimensional vector.

To solve the optimization problem \eqref{eq:reverse TRPO problem}, consider its Lagrangian given by
\begin{equation}\label{eq:lagrangian of reverse TRPO problm}
	L (\pi) = \sum_{a \in \set{A}, x \in \set{X}} \rho_k (x) \left( \pi (a|x) A^{\pi_k} (x, a) - \tau \sum_{b \in \set{A}} \pi (b|x) \log \frac{ \pi (b | x) }{ \pi_k (b | x) } \right) + \sum_{y \in \set{X}} \mu (y) \left( 1 - \sum_{b \in \set{A}} \pi (b|y) \right),
\end{equation}
where we omit terms for constraints $1 \geq \pi (a|x) \geq 0$ since the solution automatically satisfies it. Its partial derivative with respect to $\pi (a|x)$ must satisfy
\begin{equation*}
	\frac{\partial L (\pi)}{\partial \pi (a|x)} = \rho_k (x) \left( A^{\pi_k} (x, a) - \tau \log \frac{ \pi (a | x) }{ \pi_k (a | x) } - \tau \right) - \mu (x) = 0.
\end{equation*}
For $x$ such that $\rho_k (x) = 0$, any $\pi (a|x)$ is optimal. Accordingly, we may assume $\rho_k (x) \neq 0$ for all states without loss of generality. Solving for $\pi (a|x)$, we obtain
\begin{equation*}
	\pi(a|x) = \frac{\pi_k (a|x) \exp \left( A^{\pi_k} (x, a) / \tau - 1 \right)}{ \exp \left( \mu (x) / (\tau \rho_k (x)) \right) }.
\end{equation*}
From the constraint $\sum_{a \in \set{A}} \pi (a|x) = 1$, $\mu (x) = \tau \rho_k (x) \sum_{b \in \set{A}} \pi_k (b|x) \exp \left( A^{\pi_k} (x, b) / \tau - 1 \right)$. Therefore, $\pi(a|x)$ is given by
\begin{equation*}
	\pi(a|x) = \frac{\pi_k (a|x) \exp \left( A^{\pi_k} (x, a) / \tau \right)}{ Z(x; \tau) },
\end{equation*}
where $Z(x; \tau) := \sum_{b \in \set{A}} \pi_k (b|x) \exp \left( A^{\pi_k} (x, b) / \tau \right)$ is a partition function. $\beta$ is defined as $1/\tau$.

\section{GRAPE with Tables}\label{appendix:frozenlake details}

In this appendix, we describe experiment details of the experiments with value tables. Algorithm~\ref{algo:grape pseudo-algorithm} is used in both NChain and FrozenLake experiments.

\subsection{Policy Evaluation in NChain}
The experiments in NChain are conducted as follows: one trial consists of $200,000$ interactions, i.e. time steps, of an agent with an environment. At each time step, the agent takes an action $a \sim \mu (\cdot | x)$ given a current state $x$. Then, it observes a subsequent state $y$ with an immediate reward $r$. If the state transition is to a terminal state, an episode ends, and the agent starts again from a random initial state. The interactions are divided into multiple blocks. One block consists of $N = 250$ time steps. After each block, the agent update its value function using $N$ samples of the state transition data $(x, a, r, y, \mu (a|x), d)$ in the block, where $d=1$ if the transition is to a terminal state otherwise $0$. After each block, the agent is reset to the start state. $\Psi_0$ is initialized to $\Psi_0 (x, a)$. $\pi (\cdot | x)$ and $\mu (\cdot | x)$ are sampled from $|\set{A}|$-dimensional Dirichlet distribution with all concentration parameters set to $1$. The discount factor is $0.99$, and $\lambda$ is varied. A pseudo-code is shown in Algorithm~\ref{algo:grape in nchain}.

\begin{algorithm}[ht]   
	\caption{GRAPE in NChain}
	\label{algo:grape in nchain}                       
	\begin{algorithmic}
		\REQUIRE  An OpenAI/Gym-like NChain environment $env$, $\alpha \in [0, 1)$, $\lambda \in [0,1]$, and a replay buffer $\set{D}$.
		\STATE Initialize $\Psi_0 (x, a)$ to $0$ for each state $x$ and action $a$.
		\STATE Initialize $\pi (\cdot | x)$ and $\mu (\cdot |x)$ to samples from a Dirichlet distribution with all concentration parameters set to $1$ for each state $x$.
		\FOR{\text{$k$ from $0$ to $800$}}
		    \STATE $error_k \gets \sum_{x \in \set{X}, a \in \set{A}}\left( A^{\pi} (x, a) - (1 - \alpha) \Phi_k (x, a) \right)^2$.
		    \STATE $x_0 \gets env.reset()$
		    \FOR{\text{$t$ from $0$ to $249$}}
		        \STATE $a_t \sim \mu (\cdot | x_t)$, and $x_{t+1}, r_t, d_t, _ \gets env.step(a_t)$.
    		    \STATE Append $(x_t, a_t, r_t, x_{t+1}, \mu (a_t | x_t), d_t)$ to $\set{D}$.
        	\ENDFOR
    		\STATE Compute update target $\reallywidehat{\op{G^{c \mu}_\lambda} \Psi}_t$ using Algorithm~\ref{algo:grape pseudo-algorithm} and data in $\set{D}$.
    		\STATE $\Psi_{k+1} (x, a) \gets |\set{T}|^{-1} \sum_{t \in \set{T}} \reallywidehat{\op{G^{c \mu}_\lambda} \Psi}_t$, where $\set{T} := \{t | x_t = x\}$.
    		\STATE Discard samples in $\set{D}$.
		\ENDFOR
		\STATE $error_k \gets \sum_{x \in \set{X}, a \in \set{A}}\left( A^{\pi} (x, a) - (1 - \alpha) \Phi_k (x, a) \right)^2$.
		\RETURN $(error_0 / error_0, error_1 / error_0, \ldots, error_{800} / error_0)$.
	\end{algorithmic}
\end{algorithm}

\subsection{Control in FrozenLake}

The experiments in FrozenLake are done as follows: one trial consists of $5,000,000$ interactions. At each time step, the agent takes an action $a \sim \pi_k (\cdot | x)$, which is repeatedly updated through the trial, given a current state $x$. Then, it observes a subsequent state $y$ with an immediate reward $r$. If the state transition is to a terminal state, an episode ends, and the agent starts again from the start state. The state transition data $(x, a, r, y, \pi_k (a|x), d)$ are stored in a buffer $\set{D}$, of size $500,000$. Every $N = \{250, 2000\}$ (fixed through the trial) time steps, the agent updates its value function using $N$ contiguous samples from the buffer $\set{D}$. Every $100,000$ time steps, the agent updates its policy according to a rule explained below. $\beta \in \{0.1, 0.2, 0.5, 1, 2, 5, 10, 20, 50, 100\}$ are tried for each parameter set $(\alpha, \lambda, N)$ (or $(\eta, \lambda, N)$ when Retrace with a learning rate is used), and we selected one that yielded the highest asymptotic performance. $\Psi_0$ is initialized to $\Psi_0 (x, a)$. $\pi_0$ and $\mu_0$ are initialized to $\pi_0 (a|x) = \mu_0 (a|x) = 1 / |\set{A}|$. A pseudo-code is shown in Algorithm~\ref{algo:grape in frozenlake}.

\begin{algorithm}[ht]
	\caption{GRAPE in FrozenLake}
	\label{algo:grape in frozenlake}                       
	\begin{algorithmic}
		\REQUIRE  An OpenAI/Gym 8x8 FrozenLake environment $env$, $\alpha [0, 1)$, $\beta \in (0, \infty)$, $\lambda \in [0,1]$, number of steps between critic updates $N$, and a replay buffer $\set{D}$.
		\STATE Initialize $\Psi_0 (x, a)$ to $0$ for each state $x$ and action $a$.
		\STATE Initialize $\pi_0 (a | x)$ to $1 / |\set{A}|$ for each state $x$ and action $a$.
		\STATE $x_0 \gets env.reset()$
		\STATE Compute goal reaching probability $prob_0$ of $\pi_0$, and $k \gets 1$.
		\FOR{\text{$t$ from $0$ to $4,999,999$}}
	        \STATE $a_t \sim \pi_k (\cdot | x_t)$, and $x_{t+1}, r_t, d_t, _ \gets env.step(a_t)$.
    		\STATE Append $(x_t, a_t, r_t, x_{t+1}, \pi_k (a_t | x_t), d_t)$ to $\set{D}$.
    		
    		\IF{$mod(t+1, N) == 0$}
            	\STATE Compute update target $\reallywidehat{\op{G^{c \mu}_\lambda} \Psi}_u$ using Algorithm~\ref{algo:grape pseudo-algorithm} and $N$ contiguous samples from $\set{D}$.
	    		\STATE $\Psi_{k+1} (x, a) \gets |\set{T}|^{-1} \sum_{u \in \set{T}} \reallywidehat{\op{G^{c \mu}_\lambda} \Psi}_u$, where $\set{T} := \{u | x_u = x, \text{$x_u$ is in the samples.}\}$.
    		\ENDIF
    		
    		\IF{$mod(t+1, 100,000) == 0$}
            	\STATE Compute $\pi_{k+1}$ according to update rule \eqref{eq:TRPO-like policy update}.
            	\STATE Compute goal reaching probability $prob_k$ of $\pi_k$, and $k \gets k+1$.
    		\ENDIF

            \IF{$d_t == 1$}
                \STATE $x_{t+1} \gets env.reset()$
            \ENDIF

		\ENDFOR
		\RETURN $(prob_0, prob_1, \ldots, prob_{50})$.
	\end{algorithmic}
\end{algorithm}

\section{GRAPE with Neural Networks}
\label{appendix:ac}

\begin{algorithm}[t]                    
	\caption{AC-GRAPE}
	\label{algo:ac-grape}                       
	\begin{algorithmic}
		\REQUIRE  An OpenAI/Gym-like environment $env$, and a replay buffer $\set{D}$.
		\STATE Initialize $\Psi$-network parameters $\theta$.
		\STATE Initialize target $\Psi$ network parameter $\theta'$.
		\STATE Initialize policy network parameters $\phi$.
		\WHILE{task not learned}
	         \STATE $t \gets 0, x_t \gets env.reset(), done_t \gets \text{False}$.
	         \WHILE{not $done_t$}
		        \STATE $a_t \sim \pi (\cdot | x_t; \phi)$
		        \STATE $\mu_t \gets \pi (\cdot | x_t; \phi)$
		        \STATE $x_{t+1}, r_t, done_t \gets env.step(a_t)$
		        \STATE Append ($x_t, a_t, r_t, x_{t+1}, \mu_t, done_t$) to $\set{D}$
		        \STATE $t \gets t + 1$
		        \IF{$mod(step, updateFreq) == 0$}
		             \STATE $\theta, \theta', \phi \gets UpdateParamsGRAPE(\theta, \theta', \phi, \set{D})$
		        \ENDIF
			\ENDWHILE
		 \ENDWHILE
	\end{algorithmic}
\end{algorithm}

\begin{algorithm}[t]                    
	\caption{UpdateParamsGRAPE}
	\label{algo:update-params}                 
	\begin{algorithmic}
	    \REQUIRE $\Psi$-network parameters $\theta$.
	    \REQUIRE target $\Psi$ network parameter $\theta'$.
	    \REQUIRE policy network parameters $\phi$.
	    \REQUIRE a replay buffer $\set{D}$.
		\STATE Sample a trajectory $(x_t, a_t, r_t, x_{t+1}, \mu_t, d_t)$ from $\set{D}$.
		\STATE $\pi_t \gets \pi (a_t | x_t ; \phi)$.
		\STATE $\Psi_t \gets \Psi (x_t, a_t ; \theta)$.
		\STATE $\reallywidehat{A}_t (\cdot) \gets (1 - \alpha) (\Psi (x_t, \cdot ; \theta) - \pi \Psi (x_t ; \theta))$.
		\STATE Compute $\reallywidehat{\Psi_{t}}$ by $GRAPE$ with above values.
		\STATE $\mathcal{L}_\theta \gets \beta_v \sum_t \frac{\pi_t}{\mu_t} (\reallywidehat{\Psi_t} - \Psi (x_t, a_t ; \theta))^2$
		\STATE $\mathcal{L}_\phi \gets  - \beta_a \sum_t \left[\frac{\pi (a_t | x_t;\phi)}{\mu_t}  \reallywidehat{A}_t (a_t)  + Regularization\right]$.
		\STATE Update $\theta, \phi$ by using $\nabla_{\theta} \mathcal{L}_\theta$ and $\nabla_{\phi} \mathcal{L}_\phi$.
		\STATE Update $\theta'$ by $\theta' = (1-\omega) \theta' + \omega \theta$.

	\end{algorithmic}
\end{algorithm}

%\begin{algorithm}[t!]
%\SetAlgoLined
%\DontPrintSemicolon
%\KwIn{\;
%    \hspace{1em}Q-network parameters $\theta$.\;
%    \hspace{1em}target Q network parameter $\theta'$.\;
%    \hspace{1em}Policy network parameters $\phi$.\;
%    \hspace{1em}A replay buffer $\set{D}$.\;
%    }
%\;
%Sample a trajectory $(s_t, a_t, r_t, s_{t+1}, \mu_t)$ from $\set{D}$.\;
%$\pi_t \gets \pi (a_t | s_t ; \phi)$.\;
%$Q_t \gets Q (s_t, a_t ; \theta)$.\;
%$V_t \gets \pi Q (s_t ; \theta)$.\;
%$\reallywidehat{A}_t (\cdot) \gets (1 - \alpha) (Q (s_t, \cdot ; \theta) - \pi Q (s_t ; \theta))$.\;
%Compute $\reallywidehat{Q_{t}}$ by $GRAPE$ with above values.\;
%$\mathcal{L}_\theta \gets \beta_v \sum_t (\reallywidehat{Q_t} - Q (s_t, a_t ; \theta))^2$\;
%$\mathcal{L}_\phi \gets  - \beta_a \sum_{t,a} \pi (a | s_t;\phi)  \reallywidehat{A}_t (a) $ \;
%$\phantom{\mathcal{L}_\phi \gets} + \beta_h \sum_{t,a} \pi (a | s_t;\phi)　\log{\pi (a | s_t;\phi)}.$ \;
%Update $\theta, \phi$ by using $\nabla_{\theta} \mathcal{L}_\theta$ and $\nabla_{\phi} \mathcal{L}_\phi$.\;
%Update $\theta'$ by $\theta' = (1-\omega) \theta' + \omega \theta$.\;
%
%\caption{UpdateParams\ac{grape}}
%\label{algo:update-params}
%\end{algorithm}

Here we explain the experimental details of AC-GRAPE. The procedure of learning is summarized in algorihtm~\ref{algo:ac-grape}. A perceptron with 2 hidden layers was used as approximator for both the $\Psi$ function and policy function, where the first layer has 200 neurons while the second has 100. The second layer output the state-action value function via a linear layer, and the policy function via a softmax layer. We used \textit{tanh} activation for the hidden layers. A target network with the same structure was used for computing the target of $\Psi$. We applied soft-update of the target network with $\omega=0.001$, where $\omega$ is the proportion of parameter update in the target network at each training step.

We used two separate Adam optimizers for the actor and critic. The loss of actor is the mean square error of $\Psi$ function, while the loss of critic is
\begin{align}
    \mathcal{L}_\pi = - \frac{\pi}{\mu} \reallywidehat{A}  - \beta_h H(\pi), \nonumber
\end{align}
where $H$ is the entropy of policy $\pi$, and we used $\beta_h = 0.01$.

To empirically show the performance of GRAPE and Retrace across different hyper-parameters, we did a grid search on three hyper-parameters: learning rate of critic $\beta_v$, learning rate of actor $\beta_a$, and $\lambda$:
\begin{align}
    &\beta_v =  0.0001, 0.0003, 0.001, 0.003 \nonumber\\
    &\frac{\beta_a}{\beta_v} =  0.03, 0.1, 0.3, 1.0 \nonumber\\
    &\lambda =  0.0, 0.2, 0.4, 0.8, 1.0 \nonumber
\end{align}

Each experiment was repeated for 50 times, for more statistically reliable results. The best hyper-parameter setting is considered as the setting that lead to the highest overall mean reward, averaging all time steps in test phases.

Although this implementation was for discrete action space, it is possible to extend it to continuous control tasks by replacing the term $\pi \Psi$ with state value function $V$. This can be straightforwardly done by adding a function approximator of $V$, and keeping updating $V$ using $\pi$ and $\Psi$ during learning, like in \citep{wang_acer_2016}.

\end{document}